\documentclass[cco]{imc}
\usepackage{hyperref}
\usepackage{pifont}
\usepackage{multicol}

\autorNomeA{Matheus} 

\autorSobrenomeA{Martins Batista}

\autorNomeB{}

\autorSobrenomeB{}

\titulo{Análise Comparativa dos Modelos para Detecção de \textit{Deepfake}: Novas Abordagens e Perspectivas}

\orientadorNome{Isabela Neves}

\orientadorSobrenome{Drummond}

\orientadorTitulacao{Profa. Dra.}

\coorientadorNome{}

\coorientadorSobrenome{}

\coorientadorTitulacao{}

\txtOrientador{Orientador}
\txtCoorientador{Coorientador}

\ano{2024}

\mes{Novembro}

\local{Itajubá}

\palavraChaveA{Detecção de \textit{Deepfake}}
\palavraChaveB{Redes Neurais Convolucionais}
\palavraChaveC{\textit{Transformers}}
\palavraChaveD{\textit{GenConViT}}
\palavraChaveE{Benchmark de \textit{Deepfake}}

\keywordA{Deepfake Detection}
\keywordB{Convolutional Neural Networks}
\keywordC{Transformers}
\keywordD{GenConViT}
\keywordE{Deepfake Benchmark}

\membroBancaA{Profa. Dra. Isabela Neves Drummond} 
\membroBancaB{Prof. Dr. Bruno Guazzelli Batista}
\membroBancaC{M. Sc. Flávio Belizário da Silva Mota}
\membroBancaD{}
\membroBancaE{}

\dataDefesa{26 de Novembro de 2024}

\pdfAprovacao{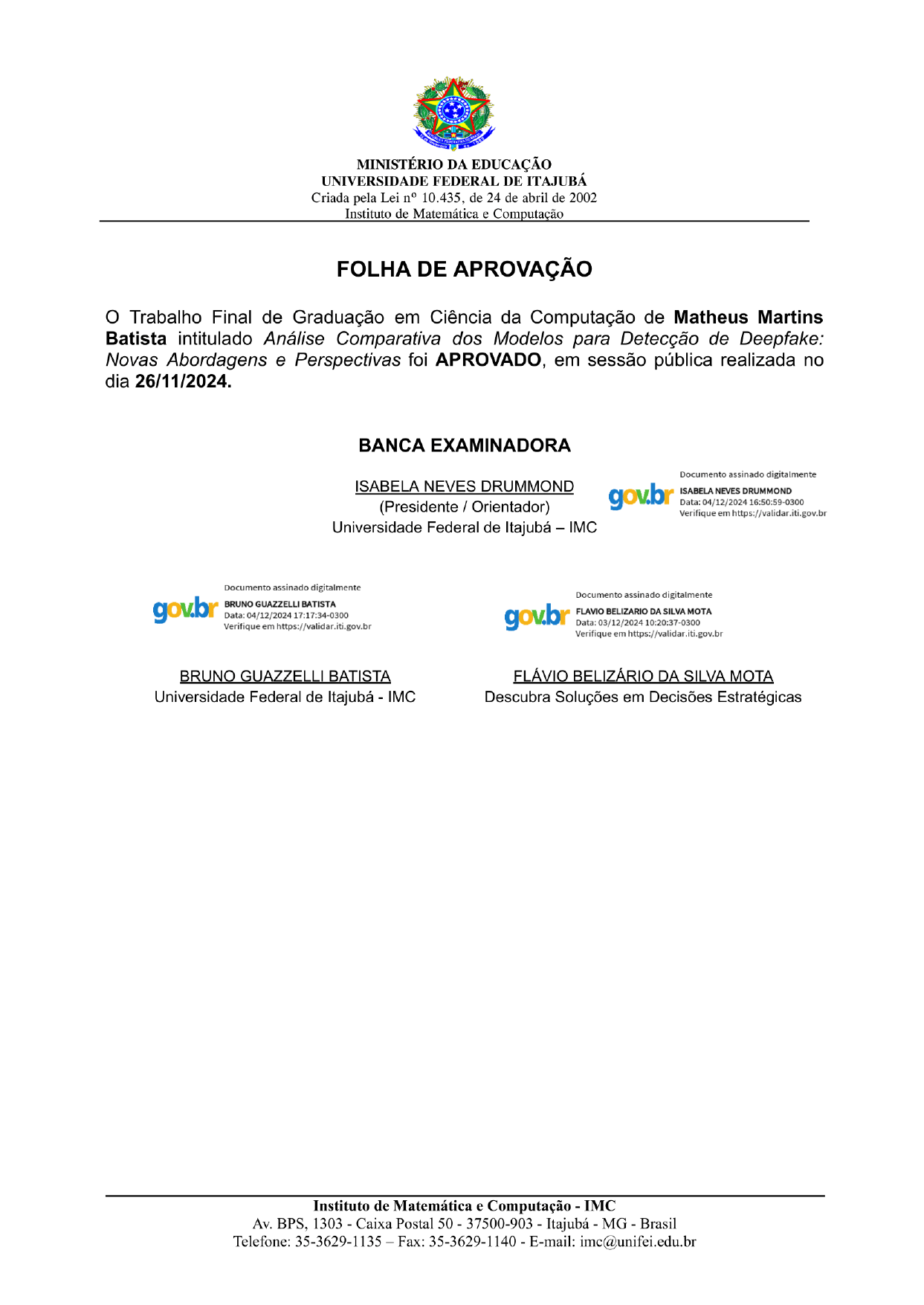} 

\begin{document} 

\frontmatter 

\maketitle 

\insereEpigrafe

\insereAgradecimentos

\insereResumo 

\insereAbstract 

\insereListaFiguras 

\insereListaTabelas 

\insereListaAbreviaturas 

\insereListaSimbolos 

\insereSumario 

\mainmatter 

\chapter{Introdução}
\label{introducao}

As tecnologias relacionadas a manipulação de mídia não são recentes e sempre atraíram a atenção do público em geral. Os sistemas de edição de fotos como o \textit{Photoshop} e \textit{CorelDRAW} mostraram-se muito úteis ao decorrer dos anos. Nesse contexto, o advento das técnicas de Aprendizado de Máquina atuais, com destaque para o Aprendizado Profundo - \textit{\ac{DL}}, permitiu a construção de uma série de sistemas intitulados como \textit{\ac{DF}}, que são capazes de manipular, adulterar e gerar imagens, vídeos e áudios que continuam a atrair a atenção das pessoas. Ademais, deve-se esse interesse tanto ao seu potencial para fins de entretenimento legítimo, quanto ao risco que representam, viabilizando a substituição de faces entre pessoas para aplicações questionáveis, como construção de farsas, notícias falsas, fraudes financeiras e pornografia falsa \citep{tolosana2020deepfakes}. Dentre os métodos de \ac{DF} que permitem criar imagens e vídeos falsos com difícil verificação de autenticidade, vale destacar o uso da técnica de \textit{\ac{GAN}} - Rede Adversária Generativa \citep{yu2021survey}.

Por uma perspectiva otimista, há inúmeras áreas de aplicação de \textit{Deepfake} para entretenimento, educação e simulações. Para \cite{mirsky2021creation}, as \textit{DeepFakes} podem ser utilizadas na dublagem de filmes estrangeiros, na educação para reanimar figuras históricas, e para experimentar acessórios em compras \textit{online}. De fato essa tecnologia tem beneficiado setores como cinema e efeitos visuais - \textit{Visual Effects} (VFX). Para ilustrar, uma cena da série \textit{The Mandalorian} foi alvo de duras críticas após tentar recriar o rosto do personagem Luke Skywalker com Computação gráfica e, com isso, o influenciador digital \textit{Shamook} que compõe o time de efeitos especiais da \textit{LucasFilm} lançou uma versão recriada com \ac{DF}, tornando-se um viral nas redes sociais \citep{bbc_shamook_2021}.

Por outro lado, se tais técnicas forem utilizadas pelas pessoas erradas, constituem ameaça para a sociedade de modo geral, uma vez que possibilitam o uso para manipulação da identidade de indivíduos, visando à geração de evidências forjadas, campanhas políticas e publicitárias maliciosas. Segundo \cite{swathi2021deepfake}, o \textit{DeepFake} pode criar vídeos pornográficos de celebridades utilizando a troca de faces ou criar discursos falsos para os líderes políticos no momento de eventos públicos. Qualquer pessoa com uma boa reputação ou tendo uma imagem pública pode ser alvo dessas atividades maliciosas. Ao analisar a origem, conforme relatado por \cite{yu2021survey}, o primeiro conteúdo \textit{DeepFake} foi postado no Reddit em 2017 por um usuário chamado "\textit{DeepFakes}". Nessa postagem havia um vídeo pornográfico de uma celebridade, expondo como é inevitável que a tecnologia seja utilizada para o mal. Esse perigo à integridade da informação pode acarretar consequências sobre a privacidade, aspectos legais, política, segurança e a erosão potencial da confiança \citep{chu2020white}.

Diante do cenário supracitado, a rápida expansão dos modelos que buscam sintetizar conteúdos falsos acarreta na necessidade de ferramentas e tecnologias para lidar com \textit{Deepfake}. A detecção de \textit{Deepfakes} tornou-se um dos grandes desafios a serem enfrentados com a expansão e o acesso público às Inteligências Artificiais. Concomitantemente, há diversas pesquisas relacionadas a esse tema na literatura com incentivos de governos e grandes indústrias. Conforme \cite{swathi2021deepfake}, ao decorrer de 2020, empresas como a Amazon, Facebook e Microsoft uniram-se para criar o \textit{\ac{DFDC}} visando construir modelos capazes de detectar imagens e vídeos falsos. Além disso, vale citar outras pesquisas sendo realizadas como: o \textit{\ac{DFGC}} ou o programa \textit{Media Forensics} da \textit{Defense Advanced Research Projects Agency} (DARPA) que também está trabalhando em tecnologias de detecção \textit{DeepFake} \citep{yu2021survey}.

Este trabalho final de graduação tem como objetivo principal investigar e comparar diferentes abordagens para a detecção de \textit{deepfakes}, contribuindo para o avanço do estado da arte nesta área crucial. Isto posto, foi conduzida uma revisão abrangente da literatura sobre detecção de \textit{deepfakes}, com foco nos principais modelos, ferramentas para extração de características, conjuntos de dados utilizados e suas respectivas métricas de desempenho. Essa revisão forneceu um panorama do estado atual da pesquisa, permitindo avaliar o desempenhos das ferramentas responsáveis pelo processos de pré-processamento e possibilitando a comparação de arquiteturas mais recentes. O \textit{GenConViT}, proposto por \cite{wodajo2023deepfake}, é um modelo híbrido que combina diferentes tipos de Redes Neurais para detectar \ac{DF}, sendo a principal arquitetura explorada neste trabalho. Buscando evidenciar os avanços obtidos, o \textit{GenConViT} também foi comparado com modelos presentes no \textit{DeepfakeBenchmark}, uma plataforma de avaliação que reúne diferentes modelos de detecção de \ac{DF}, oferecendo um ambiente padronizado para testar e comparar algoritmos de maneira justa e consistente. Ele inclui uma variedade de arquiteturas que representam o estado da arte de detecção, como \textit{XceptionNet} \citep{Afchar_2018}, \textit{EfficientNetB4}\citep{tan2019efficientnet}, \textit{Meso4Inception} \citep{Afchar_2018}, \textit{\ac{SPSL}} - \citep{liu2021spatial} e \textit{\ac{UCF}} - \citep{yan2023ucfuncoveringcommonfeatures}, que são modelos conhecidos por suas capacidades de análise de imagens e vídeos manipulados. Os resultados desse estudo permitem avaliar as capacidades, desafios e limitações das novas abordagens apresentadas em relação as propostas anteriores.

Esta monografia está organizada da seguinte forma:

\begin{itemize}
    \item O Capítulo \ref{referencial} apresenta uma revisão da literatura acerca dos principais conceitos que embasam o estudo.
    
    \item O Capítulo \ref{desenvolvimento} detalha a metodologia aplicada no desenvolvimento do estudo, abrangendo a implementação das arquiteturas de detecção, especialmente do \textit{GenConViT}, e a descrição dos conjuntos de dados utilizados. Ademais, discorre sobre o processo de ajuste fino dos modelos pré-treinados.
    
    \item O Capítulo \ref{resultados} apresenta os resultados obtidos nas comparações entre as arquiteturas do \textit{GenConViT} e os modelos do \textit{DeepfakeBenchmark}. São discutidos o desempenho de cada modelo, suas capacidades de generalização para diferentes tipos de manipulação de vídeos e as implicações dos tempos de execução nas análises.
    
    \item Por fim, o Capítulo \ref{conclusoes} ressalta os aspectos mais relevantes da técnica e as conclusões decorrentes da análise dos resultados, destacando as contribuições do estudo, as limitações identificadas e possíveis direções para trabalhos futuros na área de detecção de \ac{DF}.
\end{itemize}
\chapter{Referencial Teórico}
\label{referencial}

Neste Capítulo são discutidos os impactos das \ac{DF}s, categorizado os tipos de ataques com essa tecnologia e o processo de criação, além de apresentar um mapeamento dos domínios da detecção até presente momento. Ademais, são abordados os conceitos de processamento de imagens fundamentais para compreender o processo de manipulação e extração de informações dessas mídias. Ainda, são introduzidos tópicos relevantes dos fundamentos das áreas de Redes Neurais e Aprendizado Profundo, essenciais ao contexto de \textit{Deepfake}. Além disso, há uma Seção dedicada para as métricas que permitem avaliar o desempenho dos métodos escolhidos. Por fim, será comentado sobre o Estado da Arte de detecção de \textit{Deepfake} e os trabalhos relacionados.

\section{\textit{Deepfake} (DF)} 
Segundo \cite{yu2021survey}, a primeira tentativa de manipulação facial na literatura pode ser encontrada no icônico retrato de 1865 do presidente dos EUA, Abraham Lincoln. O desejo de manipular essas informações culminou no desenvolvimento de áreas como Computação Gráfica, ampliando as técnicas ao decorrer do anos e permitindo a edição desses tipos de mídia. Não obstante, o desenvolvimento de novos campos como da aprendizagem profunda permitiram um notável avanço no desenvolvimento de técnicas para manipulação facial.
 
Nesse contexto, o \ac{DF} surge como uma ferramenta capaz de gerar conteúdos hiper-realistas com \ac{IA}, que parecem verídicos sob a perspectiva dos olhos humanos. Essa expressão é derivada da combinação dos termos \textit{Deep Learning} - Aprendizado Profundo e \textit{Fake} - Falso, em alusão à geração de conteúdo por \ac{RNA} - \citep{mirsky2021creation}. A manipulação de imagens de humanos está entre as aplicações mais frequentes, em que a mídia sintética resultante é produto da substituição de um rosto numa imagem ou vídeo existente pelo de outra pessoa, concomitantemente, exibindo uma reencenação com conteúdo fictício \citep{Weerawardana}. Essa tecnologia têm um grande potencial para remodelar a mídia digital, enquanto as implicações causadas a sociedade também podem ser graves. Conforme \cite{Nguyen_2022}, os \ac{DF}s tornaram-se populares devido à sua qualidade em criar vídeos adulterados e facilidade de acesso à essa tecnologia no cenário atual. Uma ampla gama de usuários pode utilizar desse recurso sem precisar ter nenhum conhecimento na área.

Essa tecnologia foi disseminada em 2017, quando um usuário da rede social \textit{Reddit}, utilizando o codinome "\textit{Deepfakes}" {} \citep{Ying_2021}, utilizou um vídeo público, motores de busca de imagens e o \textit{framework} \textit{Tensorflow} \citep{tensorflow2015-whitepaper} para produzir vídeos pornográficos forjados com faces de celebridades \citep{mirsky2021creation} e publicá-los em portais de mídia social \citep{Weerawardana}. No ano seguinte, a rede social \textit{Buzzfeed} apresentou um vídeo \textit{Deepfake}, produzido com o \textit{software} \textit{FakeApp} fornecido pelo usuário do \textit{Reddit}, em que o ex-presidente Barack Obama fazia uma palestra a respeito de \ac{DF} e levantava questões concernentes a roubo de identidade, imitações e a propagação de desinformação em mídias sociais \citep{mirsky2021creation}. Segundo \cite{yu2021survey}, é notório a inspiração do método \textit{Deepfake} no trabalho "\textit{Mask R-CNN}" de \cite{he2018mask}, em que as \ac{CNN} - Redes Neurais Convolucionais foram utilizadas para gerar imagens de troca de face. Após essa aparição, uma onda de vídeos criados com troca da face começou a se espalhar pelo mundo.

Apesar da presença de um fim recreativo, tais reencenações podem acarretar constrangimento ao seu alvo. Algumas aplicações extrapolam os limites da comédia, assumindo um viés malicioso e cobrindo um espectro de ilicitudes que se estende desde fraudes financeiras até vídeos comprometedores \citep{Weerawardana}. Os impactos sociais desse tipo de aplicação podem tornar-se nefastos, do que decorre uma série de implicações legais, uma vez que infringem direitos de imagem e direitos de propriedade intelectual, culminando com prejuízos de ordem econômica e ataque à reputação.

\subsection{Categorias de Deepfake}
\begin{figure}[!ht]
\centering
\includegraphics[width=0.95\textwidth]{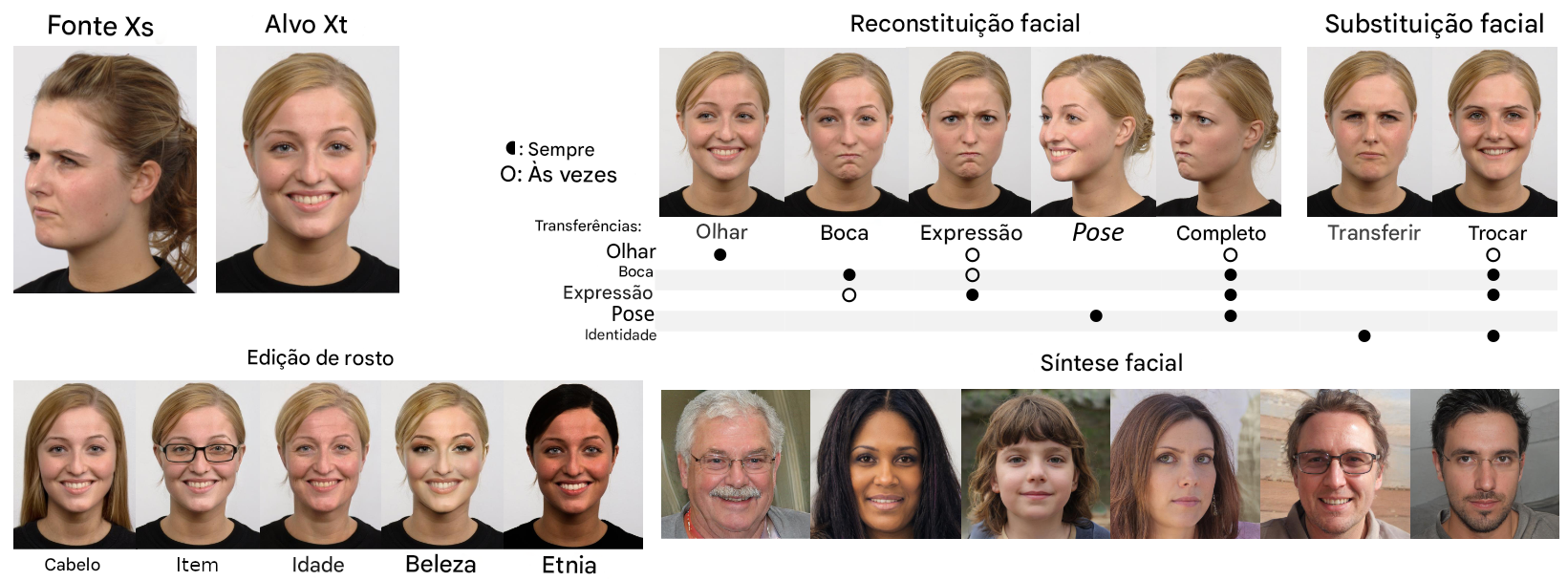}
\caption{Exemplos de \textit{Deepfake} de faces humanas por reconstituição, substituição, edição e síntese. Fonte: adaptado de \cite{mirsky2021creation}.}
\label{fig_categorias_deepfake}
\end{figure}

\cite{mirsky2021creation} apontaram para quatro categorias de \textit{Deepfake} quando o objetivo é falsificar visualmente faces humanas. Os quatro tipos identificados são a reconstituição - \textit{reenactment}, substituição - \textit{replacement}, edição - \textit{editing} e síntese - \textit{synthesis}. Para descrever cada uma delas é necessário explicar as notações por eles descritas, denotando por \textit{s} e \textit{t} as identidades de origem e alvo, respectivamente, $x_s$ e  $x_t$ como as imagens que representam aquelas identidades e $x_g$ como a imagem gerada a partir das identidades \textit{s} e \textit{t}. A Figura \ref{fig_categorias_deepfake} ilustra a categorização descrita acima.

Para a abordagem de Reconstituição, considere que a imagem de origem $x_s$ será usada para guiar a expressão, boca, olhar, posição da cabeça ou pose do corpo na imagem alvo $x_t$, conforme os detalhes presentes na Tabela \ref{tab_categoria_reconstituicao}. Essa categoria permite modelos de ataque baseados na imitação de uma identidade, controlando suas palavras para fomentar difamação, descrédito, desinformação, adulteração de evidências, falsas evidências, embaraço visando a chantagens e imitação em tempo real.

\begin{table}[hbt!]
\caption{Deepfake por Reconstituição}
\centering
\resizebox{\columnwidth}{!}{
    \begin{tabular}{|c|c|c|}
    \hline
    \rowcolor{lightgray}
    \textbf{Componente} & \textbf{Objetivo} & \textbf{Aplicações} \\ \hline
    Expressão & $x_s$ guia a expressão em $x_t$ & Cinema, \textit{video games} e mídia educacional \\ \hline
    Boca & áudio ou boca em $x_s$ guia a boca em $x_t$ & Dublagem para outro idioma e edição \\ \hline
    Olhar & direção dos olhos e globos oculares em $x_s$ guiam os correspondentes em $x_t$ & Correção de fotografias, contato visual em entrevistas \\ \hline
    Cabeça & a posição da cabeça em $x_s$ guia a correspondente em $x_t$ & Frontalização de faces \\ \hline
    Corpo & pose do corpo em $x_s$ é transferida para o corpo em $x_t$ & Síntese de pose humana \\ \hline
    \end{tabular}
}
\begin{flushleft}
\centering
\small Fonte: adaptado de \cite{mirsky2021creation}. \\
\end{flushleft}
\label{tab_categoria_reconstituicao}
\end{table}

Para a abordagem de Substituição, o conteúdo da imagem alvo $x_t$ é substituído pelo conteúdo da imagem de origem $x_s$, sob a restrição de preservar a identidade de $x_s$, conforme detalhado na Tabela \ref{tab_categoria_substituicao}. Essa categoria permite modelos de ataque como pornografia de vingança, em que a face de uma atriz é substituída pela da vítima, com fins de humilhação, difamação ou mesmo chantagem.

\begin{table}[hbt!]
\centering
\caption{Deepfake por Substituição}
\resizebox{\columnwidth}{!}{
    \begin{tabular}{|c|c|c|}
    \hline
    \rowcolor{lightgray}
    \textbf{Técnica} & \textbf{Objetivo} & \textbf{Aplicações} \\ \hline
    Transferência & Conteúdo de \(x_s\) transferido para \(x_t\) & Transferência de faces de modelos em diferentes trajes na moda \\ \hline
    Troca & Conteúdo de \(x_s\) substitui e é guiado por \(x_t\) & \textit{Face Swap} para comédia ou anonimização em vez de borramento \\ \hline
    \end{tabular}
}
\begin{flushleft}
\centering
\small Fonte: adaptado de \cite{mirsky2021creation}. \\
\end{flushleft}
\label{tab_categoria_substituicao}
\end{table}

As abordagens de Edição e Síntese representaram menores riscos de ataque se comparadas as anteriores. Apesar de uma breve descrição de cada uma presente na Tabela \ref{tab_categoria_edicao_sintese}, não foram consideradas para os estudos realizados por \cite{mirsky2021creation}. 

\begin{table}[hbt!]
\centering
\caption{Deepfake por Edição e Síntese}
\resizebox{\columnwidth}{!}{
    \begin{tabular}{|c|c|c|}
    \hline
    \rowcolor{lightgray}
    \textbf{Técnica} & \textbf{Objetivo} & \textbf{Aplicações} \\ \hline
    Edição & Alteração, remoção ou adição de atributos na imagem alvo \(x_t\) & Mudança de roupas, cabelo, idade, peso, beleza, etnia \\ \hline
    Síntese & Criação de imagem \(x_g\) sem um alvo \(x_t\) & Faces e corpos para cinema/jogos sem direitos autorais, personagens fictícios \\ \hline
    \end{tabular}
}
\begin{flushleft}
\centering
\small Fonte: adaptado de \cite{mirsky2021creation}. \\
\end{flushleft}
\label{tab_categoria_edicao_sintese}
\end{table}

\subsection{Geração de \textit{Deepfakes}}
\label{sec:geracao}

Uma vez que os vídeos \ac{DF} são o principal objeto de estudo deste trabalho, é fundamental compreender o processo de produção de imagens e vídeos (sequências de quadros) modificados ou adulterados. No cenário atual, esses conteúdos são gerados por meio de técnicas avançadas de \ac{DL}, amplamente conhecidas como \ac{IA}s generativas, que incluem \ac{GAN}s, \ac{VAE} - \textit{Autoencoders} Variacionais e Modelos de Difusão. As \ac{GAN}s, para exemplificar, consistem em duas redes neurais: um gerador \( G \), que gera dados sintéticos a partir de ruído aleatório, e um discriminador \( D \), que determina se a amostra fornecida é real ou falsa \citep{goodfellow2014generativeadversarialnetworks}. Já os \ac{VAE}s utilizam uma rede codificadora e uma decodificadora: a codificadora transforma a imagem de entrada em um espaço latente amostrado de menor dimensão, enquanto a decodificadora reconstrói uma imagem a partir desse espaço \citep{kingma2022autoencodingvariationalbayes}. As \ac{CNN}s também são utilizadas na síntese de rostos, mapeando atributos como expressão facial, idade e gênero em uma imagem facial \citep{cnn_ref}. Vale lembrar que cada uma dessas redes será melhor detalhada nas seções posteriores.

\begin{figure}[!ht]
\centering
\includegraphics[width=0.95\textwidth]{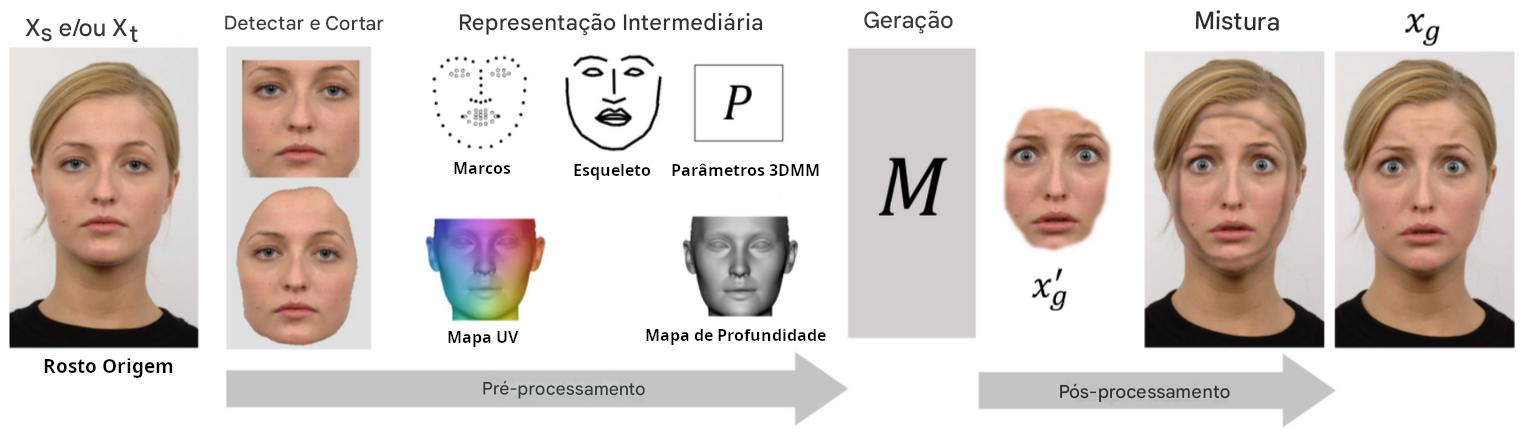}
\caption{Fluxo de Geração de Deepfake de faces humanas por reconstituição e substituição. Fonte: adaptado de \cite{mirsky2021creation}.}
\label{fig_gerar_deepfake}
\end{figure}

O processo de geração de uma imagem \ac{DF} $x_g$ por Reconstituição ou Substituição pode ser, em linhas gerais, organizado em três ou quatro etapas. Segundo \cite{Ying_2021}, esse processo inclui três estágios principais: Reconhecimento de Faces, Substituição de Faces e Pós-processamento das faces. Por outro lado, \cite{mirsky2021creation} propõem uma estrutura em quatro etapas: Detecção e Recorte da Face; Extração de Representações Intermediárias; Geração Guiada de uma Face por Outra Face; e Fusão da Face Gerada sobre a Imagem Alvo. A Figura \ref{fig_gerar_deepfake} ilustra o fluxo dessas etapas.

O primeiro estágio, a detecção e recorte da face, envolve identificar e delimitar a face na imagem, determinando seu tamanho e efetuando o recorte necessário. Esse processo é essencial para garantir que apenas a região facial seja trabalhada nas etapas subsequentes. A discussão detalhada sobre as técnicas de detecção de faces será apresentada na Seção \ref{PDI}.

Após o recorte, é realizada a extração de representações intermediárias. Nessa etapa, operações tradicionais de processamento de imagens são aplicadas à região localizada da face. Essas operações incluem conversão para tons de cinza, equalização de histograma, normalização, redução de ruído e aplicação de filtros. Em seguida, vetores de características visuais e estatísticas ou coeficientes de transformações são extraídos, representando de forma detalhada as faces. Esses vetores podem ser utilizados para consultas e comparações com padrões em bases de dados, conforme descrito por \cite{Ying_2021}.

\begin{figure}[!ht]
\centering
\includegraphics[width=0.8\textwidth]{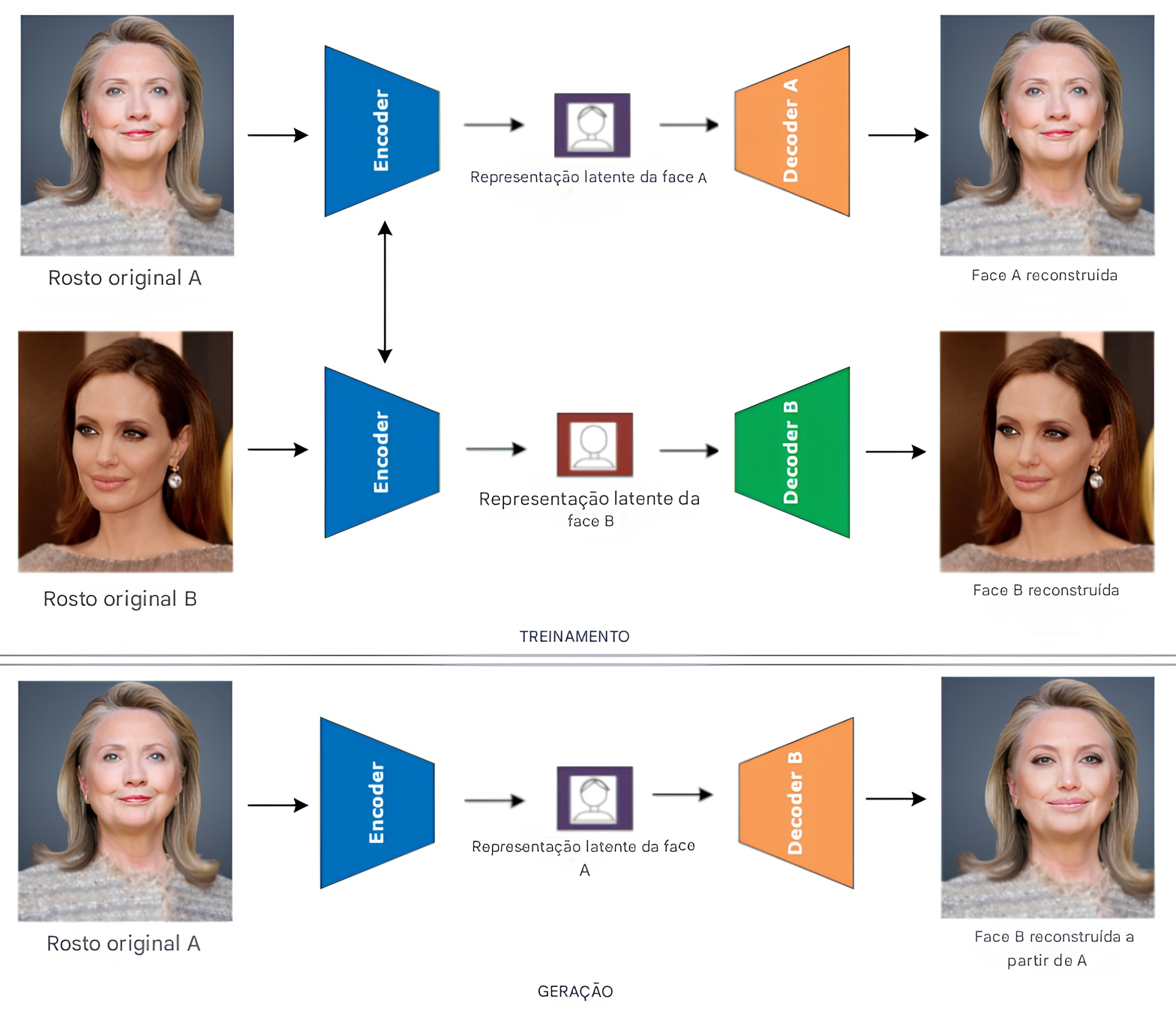}
\caption{\textit{Deepfake} gerada com \textit{autoencoder}. Fonte: adaptado de \cite{masood2021deepfakes}.}
\label{fig_gerar_com_encoder_autoencoder}
\end{figure}

A geração guiada de uma face por outra face é o próximo passo, onde uma face original é transformada na face forjada da pessoa-alvo. Esse processo frequentemente utiliza arquiteturas de \textit{Autoencoder}, conforme ilustrado na Figura \ref{fig_gerar_com_encoder_autoencoder}. O modelo de \textit{Autoencoder} é composto por uma rede codificadora (\textit{encoder}) e duas redes decodificadoras (\textit{decoder}). A rede codificadora é responsável por aprender os padrões e características comuns a todas as faces humanas, enquanto as redes decodificadoras aprendem a gerar as características específicas de cada face (original e alvo).

Concluindo as etapas, a fusão da face gerada $x'_g$ sobre a imagem alvo envolve operações de pós-processamento, que são realizadas para integrar de forma realista a face forjada à cena original. Essas operações corrigem distorções e eliminam artefatos visuais, como diferenças de tom de pele, iluminação, bordas da face e complexidade do fundo, assegurando que a face gerada seja incorporada de forma natural à imagem de fundo.

Para enriquecer o entendimento sobre os métodos de geração, vale ressaltar que as variações de \ac{GAN}s têm sido propostas com o foco em síntese facial. Algumas dessas abordagens incluem as \textit{Progressive GANs}, \textit{Wasserstein GANs} e \textit{Style-Based GANs}. As \textit{Progressive GANs} permitem a geração de imagens de alta resolução ao aumentar gradualmente a resolução das imagens durante o processo de treinamento \citep{karras2018progressive}. As \textit{Wasserstein GANs} aprimoram a estabilidade do treinamento das GANs por meio do uso de uma função de perda alternativa, que controla o processo de geração de maneira mais estável e robusta \citep{arjovsky2017wasserstein}. Já as \textit{Style-Based GANs} proporcionam controle sobre aspectos estilísticos específicos das faces geradas, como expressão facial e estilo de cabelo, permitindo ajustes mais refinados nos detalhes faciais \citep{karras2019stylebasedgan}.

Além dessas variações, há ferramentas específicas para troca facial e sincronização labial. O \textit{FaceFusion} é uma técnica que sobrepõe uma nova identidade facial ao vídeo original com ajustes de pós-processamento, como suavização de bordas e melhoria de consistência temporal \citep{ruhs2024facefusion}. Já o \textit{FaceFusion + GAN} aprimora o processo ao incluir a \textit{CodeFormer GAN}, que corrige discrepâncias visuais, especialmente em áreas como dentes e lábios, para aumentar o realismo do rosto gerado \citep{zhou2022codebook}. No contexto de \textit{deepfakes} de sincronização labial, têm-se métodos como o \textit{Wav2Lip}, que sincroniza os movimentos labiais ao áudio por meio de uma rede neural treinada em estilo \ac{GAN} \citep{prajwal2020lipsync}; e o \textit{VideoRetalking}, que aplica uma sequência de redes neurais, incluindo um modelo de reencenação guiado semanticamente, um modelo de sincronização labial e um aprimorador facial para garantir uma correspondência precisa entre o áudio e o movimento labial \citep{cheng2022videoretalking}.

\section{Processamento de Imagens Digitais}
\label{PDI}
O processamento de imagens é um método que realiza operações em uma imagem para aprimorá-la ou extrair informações relevantes \citep{Queiroz2006}. A entrada é uma imagem, e a saída pode ser uma versão aprimorada ou características associadas à imagem inicial.

Segundo \cite{pedrini2008análise}, uma imagem pode ser representada como uma função de intensidade luminosa \( f(x, y) \), onde \( x \) e \( y \) são coordenadas espaciais, e \( f(x, y) \) indica o brilho ou intensidade naquele ponto. Essa intensidade pode ser modelada pelo produto da iluminância \( i(x, y) \), que representa a luz incidente, e da reflectância \( r(x, y) \), que reflete as propriedades do objeto, conforme a Equação \ref{eq:intensidade_imagem}:

\begin{equation}
f(x, y) = i(x, y) \cdot r(x, y)
\label{eq:intensidade_imagem}
\end{equation}

O processo de digitalização transforma imagens contínuas em digitais por meio de amostragem e quantização. A amostragem discretiza o domínio espacial em uma matriz \( M \times N \), enquanto a quantização associa níveis inteiros a cada ponto da imagem. O resultado é uma matriz de \textit{pixels}, onde cada \( f(x, y) \) representa a intensidade ou cor em um ponto \citep{pedrini2008análise}. Vale salientar que as imagens podem ser monocromáticas, quando descritas apenas pela intensidade luminosa (níveis de cinza), ou coloridas, quando incluem três componentes: luminância, matiz e saturação. Essas representações são amplamente utilizadas no processamento digital para manipulação e extração de informações visuais.

\subsection{Recorte da Face}
A detecção facial, etapa crucial para o tema desse trabalho, consiste em localizar e delimitar rostos em imagens. Essa tarefa pode ser necessária no pré-processamento de conjuntos de dados ou durante predições de falsificações. Nesse cenário, podem ser utilizadas por meio de técnicas tradicionais de processamento de imagens presente na literatura, como \textit{Haar Cascade} \citep{viola2001} e \ac{HOG} - Histogramas de Gradientes Orientados combinado com \ac{SVM} - Máquina de Vetores de Suporte. No entanto, o estado da arte reside no uso de \ac{DL}, que oferece maior robustez na detecção de faces em diferentes ângulos e com oclusões. Dessa forma, além de bibliotecas já consolidadas na literatura, como a \textit{Dlib} \citep{dlib09}, que conta com modelos pré-treinados, há novas ferramentas eficientes para essa tarefa, como a \textit{Face Recognition} \citep{face_recognition} e a \textit{Deepface}.

\subsection{\textit{DeepfFace}}
A biblioteca \textit{DeepFace} \citep{serengil2024lightface} é uma ferramenta poderosa no campo do reconhecimento e análise facial. Essa tecnologia oferece uma interface simples para realizar diversas tarefas relacionadas ao processamento de imagens faciais utilizando modelos de \ac{DL} pré-treinados. Também, há uma variedade de modelos e ferramentas \textit{backend} para suportar suas funcionalidades avançadas de processamento de imagens faciais. Entre os modelos incorporados para classificação tem-se: \textit{VGG-Face, FaceNet, OpenFace, DeepFace, DeepID, ArcFace, Dlib, SFace and GhostFaceNet}. Ademais, entre as ferramentas para processar as imagens tem-se: \textit{\textbf{OpenCV}, Ssd, \textbf{Dlib}, MtCnn, Faster MtCnn, RetinaFace, \textbf{MediaPipe}, Yolo, YuNet and CenterFace}. Embora o escopo do presente trabalho não abranja todas as funcionalidades do \textit{DeepFace}, algumas bibliotecas presentes nele foram utilizadas e suas capacidades de integrar o estado da arte em detecção facial o tornaram uma referência notável.

\section{Aprendizado de Máquina}
Antes de introduzir os conceitos de \ac{DL} citados anteriormente, é fundamental definir \ac{ML} - Aprendizado de Máquina como um campo da \ac{IA} dedicado ao desenvolvimento de técnicas computacionais que permitem aos sistemas adquirir conhecimento e identificar padrões automaticamente a partir de dados, de forma semelhante ao aprendizado humano. No contexto de detecção, como a identificação de \ac{DF}, o \ac{ML} desempenha um papel essencial ao capacitar modelos a identificar padrões que indicam manipulações sutis em dados audiovisuais. Dentro do \ac{ML}, existem duas categorias principais: aprendizado supervisionado e não supervisionado \citep{Monard2003a}.

No aprendizado supervisionado, o algoritmo recebe um conjunto de exemplos com atributos de entrada e seus respectivos rótulos (atributos de saída). Esse processo é amplamente utilizado na detecção de \ac{DF}, onde o objetivo é construir um classificador capaz de distinguir entre vídeos ou imagens reais e falsos. Esse tipo de aprendizado é eficaz quando se dispõe de uma base de dados rotulada, que contém exemplos de mídia genuína e adulteradas. Se os rótulos forem categorias discretas, como "real" e "falso", o problema é chamado de classificação; se envolver previsões contínuas, ele pode se aproximar de problemas de regressão, embora a maioria das aplicações de detecção se concentre na classificação.

\begin{figure}[!ht]
\centering
\includegraphics[width=1\textwidth]{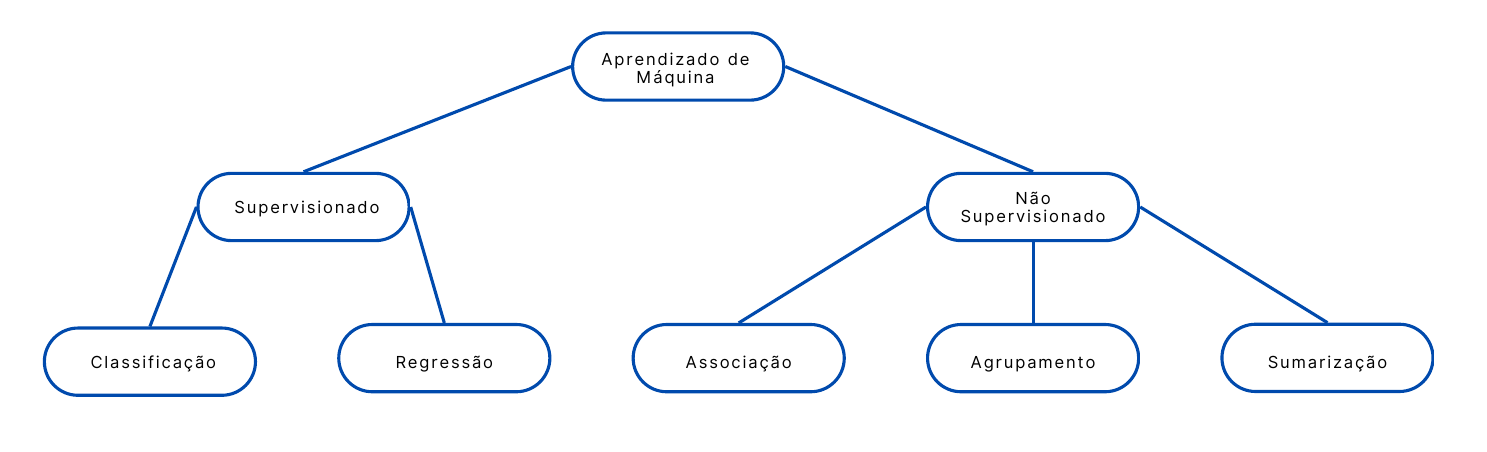}
\caption{Hierarquia de aprendizado de máquina. Fonte: Elaborado pelo autor.}
\label{fig_hierarquia_aprendizado}
\end{figure}

Por outro lado, no aprendizado não supervisionado, o algoritmo trabalha com exemplos apenas com atributos de entrada, sem rótulos, e precisa identificar padrões ou agrupamentos nos dados. Essa abordagem é útil em cenários onde os dados rotulados são escassos ou onde o objetivo é explorar características anômalas que possam indicar manipulações. Em detecção de \ac{DF}, algoritmos não supervisionados podem ser utilizados para análise de anomalias, onde o modelo aprende as características gerais de dados autênticos e identifica \textit{outliers} - pontos de desvio, que potencialmente representam \ac{DF}. A Figura \ref{fig_hierarquia_aprendizado} ilustra a hierarquia do aprendizado de máquina, mostrando a relação entre essas categorias e outras técnicas.

\section{Redes Neurais Artificiais}
\label{redes_neurais}

As \ac{RNA}s são sistemas adaptativos que aprendem a realizar tarefas a partir de exemplos, inspirados na organização das redes de neurônios biológicos. Segundo \cite{haykin2001}, uma rede neural artificial pode ser definida como um sistema de processamento de informação, cujos elementos básicos são modelos matemáticos que imitam, de forma simplificada, o funcionamento do cérebro humano. O modelo computacional das \ac{RNA}s é composto por:
\begin{itemize}
    \item \textbf{Sinapses ou elos de conexão:} São responsáveis pela comunicação entre os neurônios. Cada sinapse multiplica o sinal de entrada por um peso sináptico, que é ajustado durante o treinamento da rede.
    \item \textbf{Somador:} Combina os sinais de entrada ponderados e adiciona um viés (\textit{bias}) para ajustar a saída do neurônio, permitindo modelar deslocamentos nos dados.
    \item \textbf{Função de ativação:} Determina a saída do neurônio com base na soma ponderada dos sinais de entrada. Essa função introduz não-linearidade no modelo, o que é essencial para resolver problemas complexos. Exemplos incluem funções como \textit{sigmoid}, tangente hiperbólica (\textit{tanh}) e \textit{\ac{ReLU}}.
\end{itemize}

Esses elementos básicos trabalham em conjunto para permitir que a rede neural processe informações e aprenda padrões a partir de dados. Conforme \cite{haykin2001}, o objetivo de uma rede neural é o aprendizado de uma tarefa, ajustando seus parâmetros (pesos e viés) para minimizar a diferença entre a saída esperada e a saída obtida.

\subsection{\textit{Perceptron}}

\begin{figure}[!ht]
\centering
\includegraphics[width=0.65\textwidth]{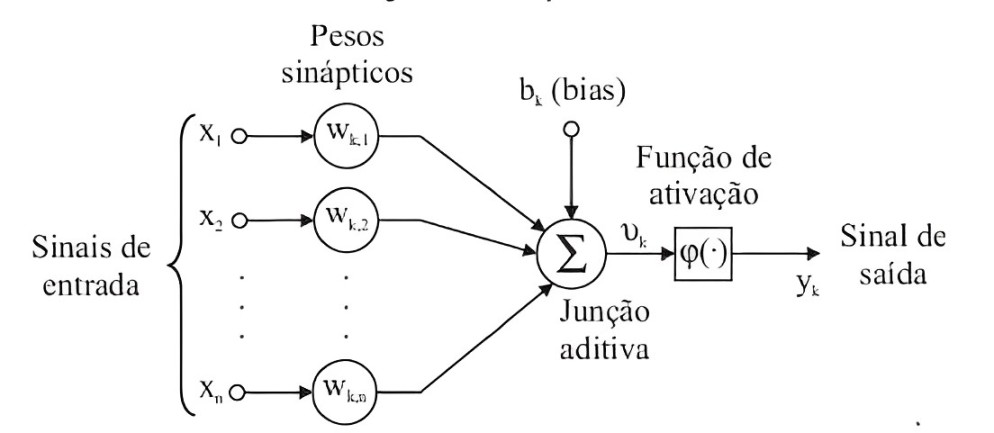}
\caption{Perceptron: representação dos sinais de entrada, pesos, soma ponderada e função de ativação. Fonte: \citep{bianchini2001}.}
\label{fig_perceptron}
\end{figure}

O Perceptron, introduzido por \cite{rosenblatt1958}, é o modelo mais simples de rede neural artificial, composto por um único neurônio. Ele foi projetado para resolver problemas de classificação linearmente separáveis, calculando o somatório ponderado das entradas (\(x_i\)) pelos pesos (\(\omega_i\)), somado a um viés (\(b\)), e aplicando uma função de ativação, tipicamente a função de degrau \citep{haykin2001}. Embora limitado a problemas linearmente separáveis e desprovido de camadas ocultas, o Perceptron foi um marco no desenvolvimento de redes neurais artificiais, servindo como base para arquiteturas mais complexas. A Figura \ref{fig_perceptron} apresenta uma representação esquemática do Perceptron, mostrando os sinais de entrada, os pesos, a soma ponderada e a aplicação da função de ativação.

\subsection{Redes Neurais Multicamadas (MLP)}

As \textit{\ac{MLP}} - Redes Neurais Multicamadas, também conhecidas como Perceptrons de Multicamadas, são uma generalização do Perceptron clássico, introduzindo uma ou mais camadas ocultas entre as camadas de entrada e saída. Cada camada é composta por múltiplos neurônios, e os neurônios de uma camada estão completamente conectados aos da próxima, formando as chamadas \textit{Fully Connected Layers (FC)} - Camadas Totalmente Conectadas \citep{vieira2017topicos}. Em constraste com Perceptron, as \ac{MLP}s podem aprender padrões complexos e resolver problemas não linearmente separáveis. Isso é possível graças ao uso de funções de ativação não lineares, como \ac{ReLU}, sigmoid ou tanh, nas camadas ocultas. 

Além disso, as \ac{MLP}s utilizam o algoritmo de retropropagação (\textit{backpropagation}) para ajustar os pesos das conexões entre os neurônios. Esse algoritmo minimiza a função de perda, atualizando os pesos de forma iterativa por meio do gradiente descendente, o que melhora a capacidade do modelo de generalizar a partir dos dados de treinamento. Essa evolução em relação ao Perceptron torna as \ac{MLP}s amplamente utilizadas em diversas tarefas de aprendizado supervisionado, como classificação, regressão e previsão de séries temporais.

\subsection{Função de Ativação}
\label{sec:func_ativicao}
A função de ativação é um componente essencial nas \ac{RNA}s, pois introduz não linearidade no modelo, permitindo que ele processe relações complexas nos dados de entrada. Conforme \cite{haykin2001}, as funções de ativação determinam a resposta dos neurônios, transformando os sinais de entrada em saídas não lineares, o que é indispensável para resolver problemas complexos e não lineares. Sem a não linearidade, as redes neurais se comportariam como simples combinações lineares, sendo incapazes de modelar problemas com alta complexidade. No presente trabalho, as principais funções de ativação discutidas são a \textit{Sigmoid}, a \textit{ReLU} e a \textit{GELU}, cada qual com propriedades e aplicações específicas.

A função \textit{Sigmoid}, conhecida também como função logística, é amplamente utilizada em redes que requerem uma saída entre 0 e 1, especialmente em classificações binárias. Ela é definida pela função expressa pela Equação \ref{eq:sigmoid}.
\begin{equation}
\sigma(x) = \frac{1}{1 + e^{-x}}
\label{eq:sigmoid}
\end{equation}

Essa função comprime a entrada em um intervalo fixo, facilitando a interpretação probabilística da saída. No entanto, ela pode sofrer com o problema do gradiente desaparecendo em redes profundas, dificultando o treinamento em camadas muito distantes da saída final \citep{glorot}.

A \textit{\ac{ReLU}} - Unidade Linear Retificada tornou-se popular devido à sua simplicidade e eficiência em redes profundas. Conforme descrito por \cite{agarap2019deeplearningusingrectified}, ela define a saída como 0 para \(x < 0\) e como uma função linear para \(x \ge 0\), sendo representada pela Equação \ref{eq:relu}.
\begin{equation}
\text{ReLU}(x) = \max(0, x)
\label{eq:relu}
\end{equation}

Esse comportamento ajuda a mitigar o problema do gradiente desaparecendo porque a ReLU mantém um gradiente constante para valores positivos de \(x\). Diferentemente de funções como \textit{sigmoid} ou \textit{tanh}, cujos gradientes diminuem exponencialmente conforme os valores de entrada se afastam do zero (tendendo a zero para entradas muito grandes ou muito pequenas), a ReLU preserva um gradiente igual a 1 para \(x \geq 0\). Isso reduz significativamente o risco de saturação, onde os gradientes se tornam tão pequenos que a atualização dos pesos é quase inexistente. Além disso, a simplicidade computacional da ReLU (ativação linear para \(x \geq 0\) e zero para \(x < 0\)) permite cálculos mais rápidos e eficientes, acelerando o treinamento da rede. Essa característica é particularmente vantajosa em redes profundas, onde o acúmulo de gradientes pequenos pode ser um problema significativo.

A função \textit{\ac{GELU}} é uma função de ativação mais recente que aplica uma ativação suave com base nas propriedades probabilísticas da entrada. Ao contrário da \ac{ReLU}, que simplesmente zera os valores negativos, a \ac{GELU} utiliza uma transição gradual, permitindo que valores negativos sejam parcialmente transmitidos, o que pode capturar melhor a informação em modelos complexos como os \textit{Transformers} \citep{hendrycks2023gaussianerrorlinearunits}. A \ac{GELU} é definida pela função \ref{eq:gelu}:
\begin{equation}
\text{GELU}(x) = x \cdot \frac{1}{2} \left( 1 + \text{erf}\left(\frac{x}{\sqrt{2}}\right)\right)
\label{eq:gelu}
\end{equation}
onde \( \text{erf}(x) \) representa a função de erro. Essa característica permite ao \ac{GELU} introduzir uma ativação contínua que se adapta à magnitude da entrada, sendo útil em tarefas que requerem um alto grau de sensibilidade e precisão.

\subsection{Retropropagação e Gradiente Descendente}

As \ac{RNA}s ajustam os pesos de cada camada para aprender a partir dos dados de entrada. A retropropagação, combinada com o gradiente descendente, é um dos algoritmos mais comuns para realizar esses ajustes, conforme destacado por \cite{vieira2017topicos}. Esse processo ocorre em duas etapas principais: o cálculo da perda na saída da rede e a propagação dos gradientes dessa perda para ajustar os pesos em todas as camadas. \cite{bianchini2001} explica que a retropropagação é um método supervisionado que requer uma resposta desejada (valor alvo). Durante o treinamento, a perda é calculada comparando o valor de saída da rede (\(x'\)) com o valor alvo (\(x\)), e o algoritmo ajusta os pesos para reduzir essa perda iterativamente. Além disso, o gradiente descendente, conforme \cite{haykin2001}, é o método utilizado para minimizar a função de perda. Ele ajusta os pesos na direção oposta ao gradiente da função de perda, de acordo com uma taxa de aprendizado. Esse processo garante que a rede se torne mais precisa ao longo do treinamento, minimizando gradativamente a diferença entre as previsões e os valores reais.

Se não controlado adequadamente, o processo de retropropagação pode levar a problemas como \textit{overfitting} ou \textit{underfitting}. \textit{Overfitting} ocorre quando o modelo se ajusta excessivamente aos dados de treinamento, resultando em baixa generalização e alto erro em novos dados. Já o \textit{underfitting} ocorre quando o modelo não consegue capturar a complexidade dos dados, apresentando alto erro tanto no treinamento quanto na validação. Técnicas como regularização (\textit{dropout}, L1/L2), ajuste de taxa de aprendizado, aumentação de dados e parada antecipada são comumente usadas para mitigar esses problemas.

\subsection{Funções de Perda}

As funções de perda desempenham um papel central no aprendizado supervisionado, pois determinam como a discrepância entre as saídas previstas pela rede (\(x'\)) e os valores reais ou desejados (\(x\)) será mensurada. Segundo \cite{Goodfellow2016}, elas fornecem uma métrica para ajustar os pesos das redes neurais durante o treinamento, permitindo que o modelo aprenda ao minimizar essa discrepância. Duas das funções mais utilizadas são o \ac{MSE} - Erro Quadrático Médio e a \ac{CE} - Entropia Cruzada, escolhidas de acordo com a natureza do problema e dos dados.

O \ac{MSE} é aplicado principalmente em problemas de regressão ou tarefas com saídas contínuas. Ele mede a média das diferenças quadráticas entre as previsões (\(x'\)) e os valores reais (\(x\)), penalizando previsões distantes. Embora amplamente utilizado, o \ac{MSE} pode ser sensível a \textit{outliers} devido à penalização quadrática. Sua fórmula é apresentada na Equação \ref{eq:mse}.

\begin{equation}
\label{eq:mse}
L_{MSE} = \frac{1}{n} \sum_{i=1}^{n} (x_i - x_i')^2
\end{equation}

A \ac{CE}, por sua vez, é amplamente empregada em problemas de classificação, especialmente quando as saídas são representadas como distribuições de probabilidade. Ela mede a divergência entre a distribuição verdadeira (\(x\)) e a prevista (\(x'\)), assumindo que \(x'\) está no intervalo \([0, 1]\). Segundo \cite{Goodfellow2016}, a \ac{CE} é comumente utilizada em conjunto com funções de ativação como \textit{sigmoid} (para classificação binária) ou \textit{softmax} (para classificação multiclasse), pois essas funções interpretam as saídas como probabilidades. Sua definição é dada pela Equação \ref{eqCE}.

\begin{equation}
\label{eqCE}
L_{CE} = - \frac{1}{n} \sum_{i=1}^{n} \left[ x_i \log(x_i') + (1 - x_i) \log(1 - x_i') \right]
\end{equation}

Essas funções de perda fornecem o sinal de erro que será retropropagado pela rede para ajustar os pesos por meio do gradiente descendente. A escolha apropriada da função de perda é crucial para o desempenho e a capacidade do modelo de generalizar para novos dados.

\section{Redes Neurais Profundas (DNN)}

As \ac{DNN} - Redes Neurais Profundas constituem uma classe de modelos de aprendizado de máquina que permitem alcançar um alto grau de exatidão em problemas complexos. Com o advento do \ac{DL} - Aprendizado Profundo, essas redes se tornaram ferramentas poderosas para lidar com tarefas que envolvem grande volume de dados, como reconhecimento de imagem, áudio, processamento de linguagem natural e muitas outras \citep{vieira2017topicos}. A \ac{DL} surgiu como uma solução para os desafios enfrentados por arquiteturas tradicionais, como \ac{RNA}s rasas e \ac{SVM}s, em contextos de alta dimensionalidade. Segundo \cite{Ludovic2011}, a alta dimensionalidade refere-se ao número de variáveis ou características utilizadas para descrever os dados, o que aumenta a complexidade do problema e torna os dados mais dispersos no espaço.

Nesta seção, serão explorados os principais tipos de \ac{DNN}s, suas características e aplicações relacionadas à detecção de \ac{DF}. Serão abordadas: as Redes Neurais Convolucionais, amplamente utilizadas para extração de características espaciais em imagens; as Redes Adversárias Generativas, amplamente conhecidas por sua capacidade de criar amostras realistas e detectar padrões falsificados; as Redes Neurais Recorrentes, projetadas para processar sequências de dados temporais; as Redes Codificadoras-Descodificadoras, que combinam codificação e reconstrução de dados; os \textit{Autoencoders} e \textit{Variational Autoencoders}, usados para compressão e geração de dados; e, finalmente, os Modelos Baseados em \textit{Transformers}, que vêm se destacando em tarefas relacionadas a aprendizado de sequência e atenção contextual.

\subsection{Redes Neurais Convolucionais (CNN)}
As \ac{CNN}s são provavelmente os modelos de \ac{DL} mais conhecidos e citados na literatura para detecção de \ac{DF}. Esse tipo de rede neural utiliza camadas convolucionais para processar a entrada, considerando campos receptivos locais. A convolução é uma parte fundamental da \ac{CNN}. Segundo \cite{Goodfellow2016}, ela é dividida em três partes: a convolução de cada camada de entrada, a aplicação de uma função de ativação não linear e uma função de subamostragem (\textit{pooling}). A etapa de convolução caracteriza-se pela passagem da máscara (\textit{kernel}) - filtro pela imagem de entrada, e o resultado desse processamento é denominado mapa de características da saída, que permite o reconhecimento de padrões. Dentre as características que diferem essas redes das demais estão: conectividade esparsa das unidades de processamento, campo receptivo local, arranjo espacial, compartilhamento de pesos e \textit{dropout}.

\begin{figure}[!ht]
\centering
\includegraphics[width=0.95\textwidth]{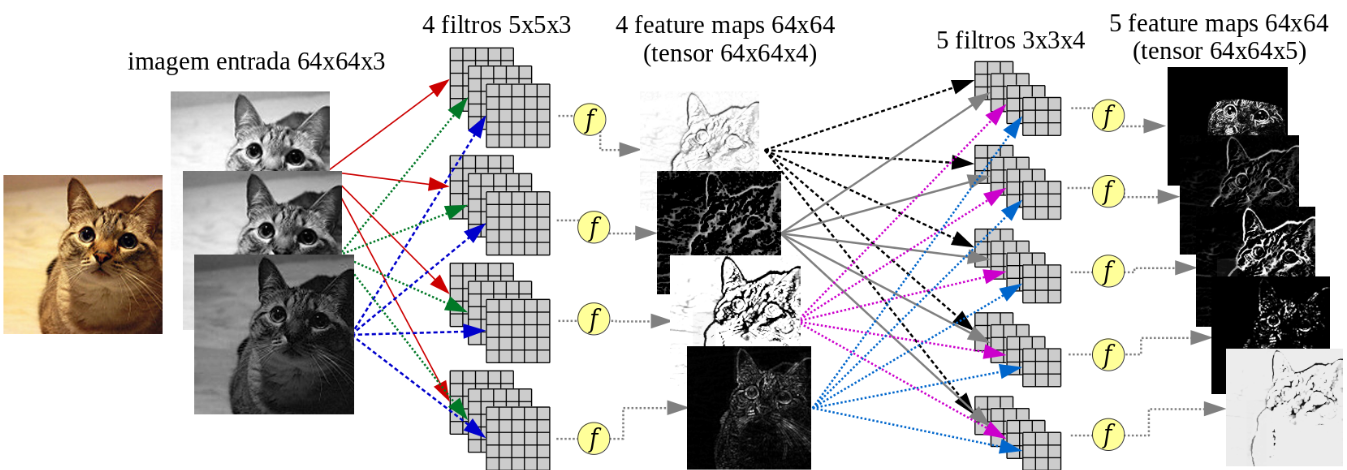}
\caption{Ilustração de duas camadas convolucionais: a primeira com 4 filtros 5 × 5 × 3, que recebe como entrada uma imagem RGB 64 × 64 × 3, e produz um tensor com 4 mapas de características; a segunda camada convolucional contém 5 filtros 3 × 3 × 4 que filtram o tensor da camada anterior, produzindo um novo tensor de mapas de características com tamanho 64 × 64 × 5. Os círculos após cada filtro denotam funções de ativação, como a \textit{ReLU}. Fonte: \citep{vieira2017topicos}.}
\label{fig_exemplo_cnn}
\end{figure}

Uma \ac{CNN} possui algumas camadas, incluindo um ou mais planos. A imagem é inserida na primeira camada, onde cada filtro (neurônio) dessa camada processa a imagem e a transforma por meio de combinações lineares com \textit{pixels} vizinhos. Vários filtros são aplicados em múltiplas camadas para que as características possam ser extraídas. A Figura \ref{fig_exemplo_cnn} ilustra um exemplo de camadas convolucionais.

Segundo \cite{Goodfellow2016}, a convolução é um operador linear que recebe duas funções como entrada e retorna uma terceira função. No contexto de dados discretos, a convolução é utilizada para combinar uma entrada (\(f\)) com um filtro ou kernel (\(g\)) para produzir um mapa de características (\(h\)). A fórmula adaptada da convolução discreta é apresentada na Equação \ref{eq:convolucao}.

\begin{equation}
(f * g)(k) = h(k) = \sum_{i=0}^{n} f(i) \cdot g(k-i)
\label{eq:convolucao}
\end{equation}
onde \( f \) e \( g \) são sequências numéricas. Essa operação é amplamente utilizada em redes neurais convolucionais para detectar padrões locais, como bordas e texturas, em dados estruturados como imagens. Os \textit{kernels} podem também ser considerados filtros de convolução que percorrem a imagem e calculam o produto escalar, onde cada filtro extrai diferentes recursos da imagem. A camada convolucional de uma \ac{RNA} é responsável por extrair as características de entrada \citep{vieira2017topicos}.

\subsection{Redes Adversárias Generativas (GANs)}

Uma \ac{GAN} é um \textit{framework} de treinamento composto por duas redes neurais: a Geradora (G) e a Discriminadora (D), treinadas de maneira adversarial. A rede G gera amostras falsas \(x_g\) para enganar a rede D, que, por sua vez, aprende a distinguir entre amostras reais (\(x \in \mathcal{X}\)) e amostras falsas (\(x_g = G(z)\)), onde \(z \sim N\) é um vetor amostrado aleatoriamente de uma distribuição normal \citep{mirsky2021creation}. 

Durante o treinamento, a rede D busca maximizar a média do logaritmo das probabilidades de imagens reais e do logaritmo das probabilidades invertidas de imagens falsas geradas, como mostrado na Equação \ref{eq:loss_D}. Simultaneamente, a rede G busca minimizar o logaritmo das probabilidades invertidas da predição de D sobre imagens geradas falsas. No entanto, essa abordagem pode levar à saturação dos gradientes quando a rede D é muito eficaz. Para superar esse problema, utiliza-se uma função de perda alternativa, em que G maximiza a probabilidade de D classificar corretamente as imagens geradas falsas, conforme mostrado na Equação \ref{eq:loss_G} \citep{goodfellow2014generativeadversarialnetworks}.

\begin{equation}
L_{adv}(D) = \max \left[ \log D(x) + \log (1 - D(G(z))) \right] 
\label{eq:loss_D}
\end{equation}

\begin{equation}
L_{adv}(G) = \min \left[ \log (1 - D(G(z))) \right] 
\label{eq:loss_G}
\end{equation}

\begin{figure}[!ht]
\centering
\includegraphics[width=1\textwidth]{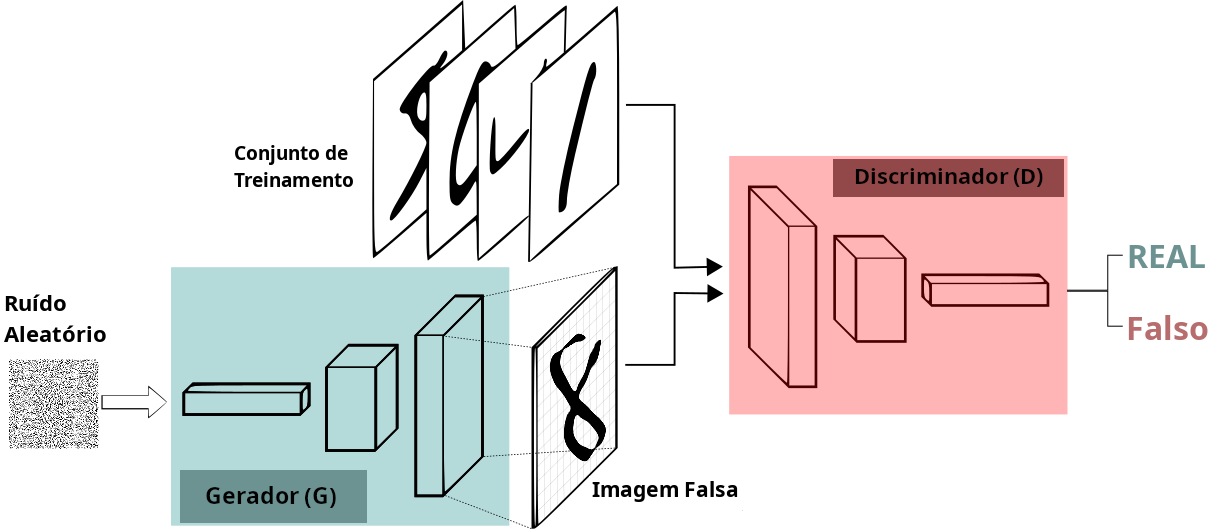}
\caption{Arquitetura de uma rede GAN. Fonte: Adaptado de \cite{Silva2018gans}.}
\label{fig_gans}
\end{figure}

Segundo \cite{santos2020}, as GANs são fundamentadas em uma arquitetura projetada para construir modelos generativos que tentam aprender a distribuição dos dados. De acordo com \cite{goodfellow2014generativeadversarialnetworks}, a dinâmica entre as redes funciona como um jogo de \textit{minimax} de dois jogadores. Enquanto a rede Geradora cria amostras que tentam enganar a Discriminadora, esta última se aperfeiçoa para distinguir cada vez melhor as amostras reais das falsas. Esse processo iterativo permite que a rede G produza amostras cada vez mais realistas. A Figura \ref{fig_gans} ilustra esse conceito. 

As GANs possuem uma ampla gama de aplicações, como geração de imagens realistas a partir de descrições textuais, melhoria da qualidade de imagens e criação de \ac{DF}. Nesse perspectiva, o \textit{framework pix2pix} \citep{Isola2017pix2pix} viabiliza transformações pareadas entre domínios. Durante o treinamento, a rede G gera uma imagem \(x_g\) dada uma imagem de contexto visual \(x_c\), enquanto a rede D discrimina entre (\(x, x_c\)) e (\(x_g, x_c\)). A rede G é uma \textit{U-Net}, uma arquitetura de \textit{Encoder-Decoder} com \ac{CNN} (ED CNN) e conexões residuais entre o \textit{Encoder} e o \textit{Decoder} para preservar detalhes da entrada, permitindo a geração de conteúdo em alta resolução \citep{mirsky2021creation}.

O \textit{framework CycleGAN} é uma evolução do \textit{pix2pix}, permitindo a tradução de imagens não pareadas. Duas \ac{GAN}s atuam em um ciclo, convertendo imagens entre os domínios de entrada e saída, encorajando a consistência por meio da função de perda de consistência de ciclo (\(L_{cyc}\)). Além dessas, as \textit{StyleGANs}, previamente discutida na Seção \ref{sec:geracao}, introduzem um controle refinado sobre o estilo e características visuais das imagens geradas, sendo amplamente utilizadas na criação de faces sintéticas e outras tarefas avançadas de geração de imagens \citep{karras2019stylebasedgan}.

\subsection{Redes Neurais Recorrentes (RNNs)}
As \ac{RNN} são modelos projetados para lidar com dados sequenciais, permitindo incorporar informações de etapas anteriores por meio de sua memória interna \citep{Goodfellow2016}. Essa característica é fundamental para tarefas como geração, classificação e tradução de sequências, especialmente em dados como vídeos e áudios, onde o contexto temporal é essencial.
Embora eficazes, as \ac{RNN}s enfrentam desafios, como o gradiente desvanecente ou explosivo, que dificultam o aprendizado em sequências longas. Para superar essas limitações, surgiram variações como a \ac{LSTM}, que utilizam células de memória e portas (entrada, saída e esquecimento) para gerenciar o fluxo de informações ao longo do tempo \citep{hochreiter1997long}, e a \ac{GRU}, que simplificam as \ac{LSTM}s com menor complexidade computacional e desempenho competitivo \citep{cho2014learning}. Na detecção de deepfakes, RNNs são aplicadas para identificar inconsistências temporais em vídeos, analisando padrões em movimentos faciais ou alterações sonoras, contribuindo para a detecção de manipulações sutis e incoerentes \citep{sabir2019recurrentconvolutionalnetworks}.

\subsection{Redes Codificadoras-Decodificadoras (ED)}

As redes \ac{ED} - Codificadoras-Decodificadoras constituem uma arquitetura composta por duas partes principais: um codificador (\textit{encoder}) \(g_\phi\), responsável por transformar os dados de entrada \(x \in \mathcal{X}\) em uma representação latente \(z\), e um decodificador (\textit{decoder}) \(f_\theta\), cuja função é reconstruir os dados de entrada a partir dessa representação latente. Essa arquitetura é amplamente aplicada em problemas que envolvem redução de dimensionalidade, geração de dados sintéticos e aprendizado não supervisionado. O objetivo geral de uma rede ED é reconstruir os dados de entrada de maneira fiel, conforme $f_\theta(g_\phi(x)) \approx x$.

\subsubsection{\textit{Autoencoders} (AE)}

Os \textit{\ac{AE}s} representam uma classe de redes \ac{ED} projetadas para aprender uma função identidade de maneira não supervisionada, reconstruindo o dado de entrada \(x\) enquanto comprimem a informação no processo, com o objetivo de descobrir uma representação mais eficiente e compacta. A ideia foi introduzida na década de 1980 e posteriormente promovida pelo artigo seminal de \cite{HintonSalakhutdinov2006}.

Os \ac{AE}s consistem em duas partes principais: Um codificador (\(g_\phi\)) que transforma os dados de entrada \(x \in \mathcal{X}\) em uma representação comprimida \(z\), também conhecida como espaço latente. Um decodificador (\(f_\theta\)) que reconstrói o dado de entrada \(x'\) a partir da representação comprimida \(z\). Essas redes são amplamente utilizadas em tarefas como compressão de dados, remoção de ruídos e aprendizado de características latentes.

\begin{figure}[!ht]
\centering
\includegraphics[width=1\textwidth]{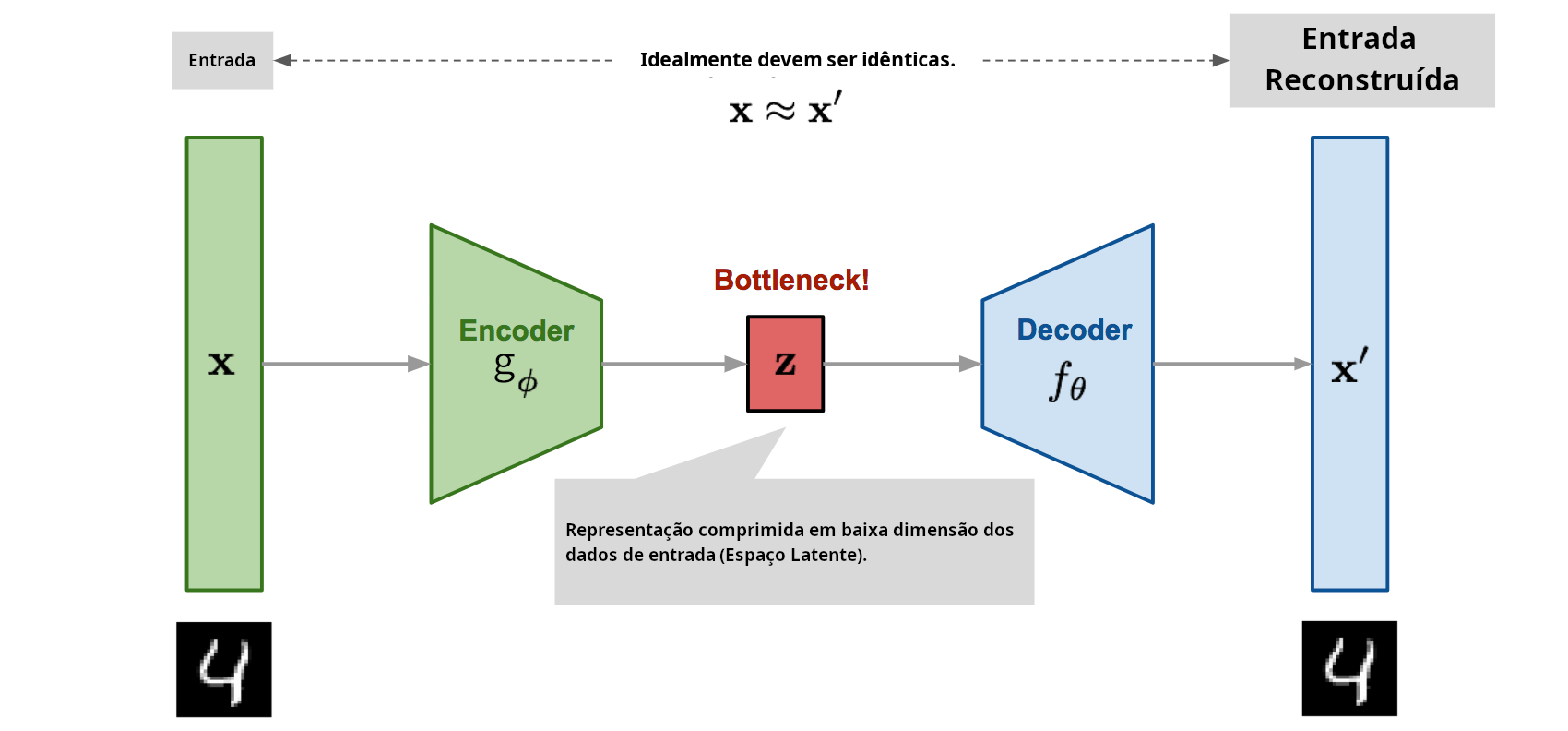}
\caption{Arquitetura de um \textit{Autoencoder}. O codificador \(g_\phi\) transforma os dados de entrada \(x\) em uma representação comprimida \(z\), também conhecida como espaço latente ou \textit{bottleneck}. O decodificador \(f_\theta\) utiliza essa representação para reconstruir os dados de entrada \(x'\). Idealmente, \(x'\) deve ser uma aproximação precisa de \(x\). Fonte: adaptado de \cite{weng2018VAE}.}
\label{fig_autoencoder}
\end{figure}

No entanto, uma limitação significativa dos \ac{AE}s tradicionais reside na possibilidade de o espaço latente gerado ser disperso e descontinuado, o que dificulta sua aplicação em tarefas que demandam geração de dados ou manipulação de vetores latentes. Em particular, os \ac{AE}s não garantem que o espaço latente seja contínuo ou que representações interpoladas entre dois pontos latentes gerem amostras válidas no domínio de entrada \citep{Goodfellow2016}. A Figura~\ref{fig_autoencoder} apresenta a arquitetura de um \textit{Autoencoder} típico.

Nos \ac{AE}s, a escolha da função de perda depende do tipo de dado que está sendo modelado. Para dados contínuos, o \ac{MSE} é frequentemente utilizado. No entanto, quando os dados de entrada são binários ou categóricos, a \ac{CE} é mais apropriada. Essa métrica mede a divergência entre distribuições probabilísticas, sendo especialmente útil quando a saída do decodificador é ativada por uma \textit{sigmoid}, produzindo valores no intervalo \([0, 1]\). A função de perda \ac{CE} é definida pela Equação \ref{eqCE}.

\subsubsection{\textit{Variational Autoencoders} (VAE)}

\begin{figure}[!ht]
\centering
\includegraphics[width=1\textwidth]{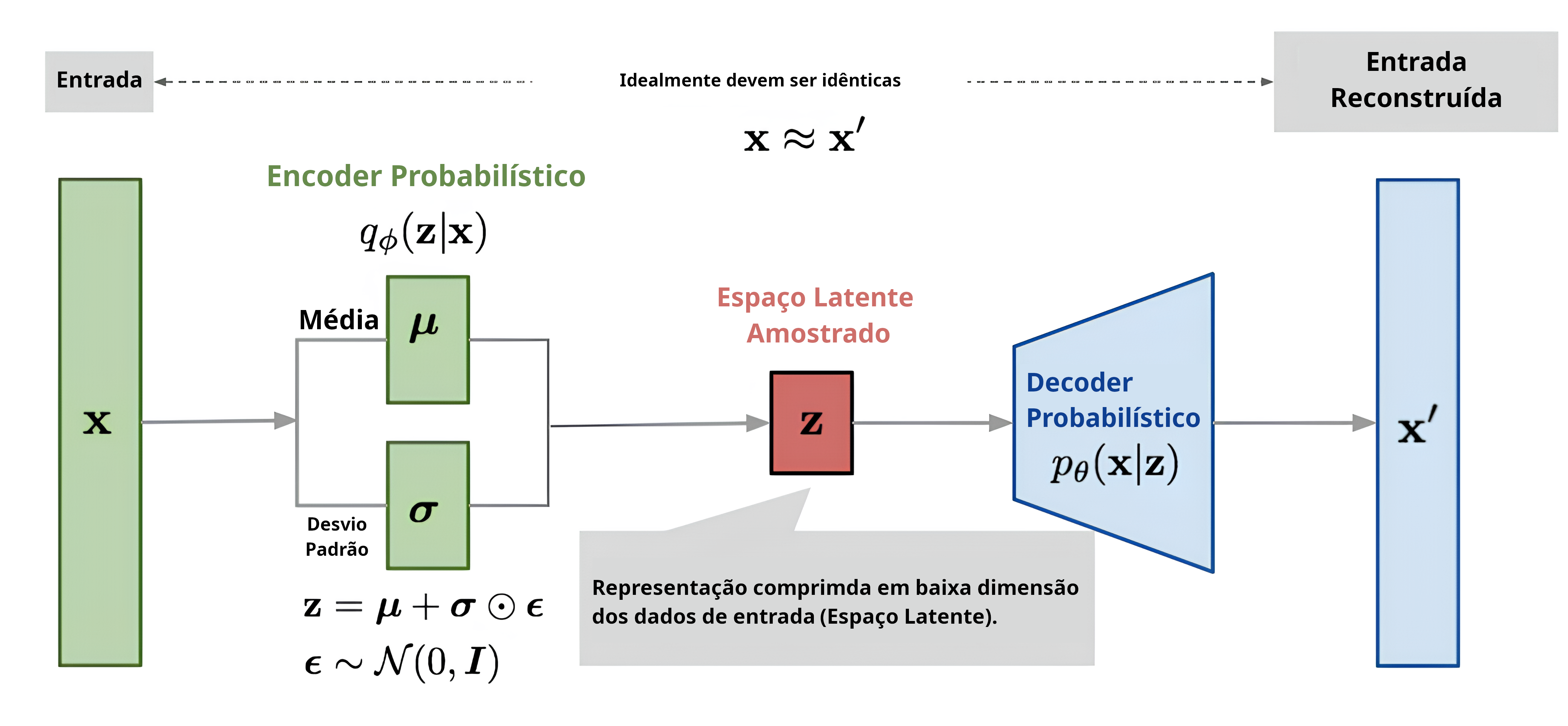}
\caption{Arquitetura de um Variational Autoencoder. O codificador \(q_\phi(z|x)\) aprende os parâmetros de uma distribuição latente (\(\mu\) e \(\sigma\)), permitindo a amostragem de vetores \(z\). Esses vetores são utilizados pelo decodificador \(p_\theta(x'|z)\) para reconstruir os dados de entrada. Fonte: adaptado de \cite{weng2018VAE}.}
\label{fig_vae}
\end{figure}

Os \ac{VAE}s representam uma extensão probabilística dos \ac{AE}s, incorporando a modelagem de distribuições no espaço latente. Diferentemente dos \ac{AE}s tradicionais, que aprendem representações determinísticas, os \ac{VAE}s aprendem uma distribuição probabilística que captura as variabilidades intrínsecas dos dados. Em vez de gerar diretamente um vetor latente \(z\), o codificador probabilístico \(q_\phi(z|x)\) estima os parâmetros (\(\mu\) e \(\sigma^2\)) de uma distribuição gaussiana, da qual \(z\) é amostrado \citep{kingma2022autoencodingvariationalbayes}. Conforme ilustrado na Figura~\ref{fig_vae}, o codificador \(q_\phi(z|x)\) transforma os dados de entrada \(x\) em uma distribuição no espaço latente parametrizada por \(\mu\) e \(\sigma\). A variável latente \(z\) é então amostrada dessa distribuição, conforme a Equação \ref{latente}:

\begin{equation}
\label{latente}
z = \mu + \sigma \odot \epsilon, \quad \epsilon \sim \mathcal{N}(0, I)
\end{equation}

O decodificador probabilístico \(p_\theta(x'|z)\) utiliza amostras \(z\) dessa distribuição para reconstruir os dados de entrada \(x'\), conforme a Equação \ref{decoder}:

\begin{equation}
\label{decoder}
x' \sim p_\theta(x'|z)
\end{equation}

A introdução do espaço latente probabilístico nos \ac{VAE}s supera as limitações dos \ac{AE}s tradicionais, como a falta de continuidade no espaço latente. Enquanto os \ac{AE}s geram vetores latentes fixos, os \ac{VAE}s modelam uma distribuição de \(z\), permitindo maior generalização, interpolação suave entre amostras latentes e geração de dados sintéticos válidos.

A função de perda dos \ac{VAE}s é composta por dois termos principais. O primeiro é o Erro de Reconstrução (\(L_{recon}\)), que mede a diferença entre os dados de entrada \(x\) e sua reconstrução \(x'\). Comumente, utiliza-se a log-verossimilhança, conforme representada pela Equação~\ref{eq:log_likelihood}, ou \ac{MSE}, apresentado na Equação~\ref{eq:mse}. Logo, tem-se a igualdade $L_{recon} = L_{MSE}$ ou $L_{recon} = L_{logv}$.

\begin{equation}
\label{eq:log_likelihood}
L_{logv} = -\mathbb{E}_{q_\phi(z|x)}[\log p_\theta(x'|z)]
\end{equation}

A Equação~\ref{eq:log_likelihood} é frequentemente utilizada para modelar distribuições probabilísticas em tarefas onde \(p_\theta(x'|z)\) representa uma distribuição condicional, como uma Gaussiana ou Bernoulli. Já a Equação~\ref{eq:mse} é mais comum em implementações onde a reconstrução \(x'\) é tratada como uma saída determinística.

O segundo termo é a Divergência de Kullback-Leibler (\(L_{KL}\)), que mede a diferença entre a distribuição latente aprendida \(q_\phi(z|x)\) e a distribuição prior \(p(z)\), geralmente uma gaussiana padrão \(\mathcal{N}(0, I)\): 

\begin{equation}
L_{KL} = D_{KL}(q_\phi(z|x) \| p(z)) = \int q_\phi(z|x) \log \frac{q_\phi(z|x)}{p(z)} dz
\end{equation}

A combinação desses dois termos define a função de perda total dos \ac{VAE}s:

\begin{equation}
L_{VAE} = L_{recon} + L_{KL}
\end{equation}

Para que o treinamento do modelo seja possível por meio de retropropagação, utiliza-se o truque de reparametrização, que reformula a amostragem estocástica \(z \sim \mathcal{N}(\mu, \sigma^2)\) como uma operação determinística: 

\begin{equation}
z = \mu + \sigma \odot \epsilon, \quad \epsilon \sim \mathcal{N}(0, I)
\end{equation}

Esse truque permite que os parâmetros \(\mu\) e \(\sigma\) sejam aprendidos diretamente durante o treinamento, enquanto a estocasticidade permanece no termo \(\epsilon\), garantindo a possibilidade de otimização do modelo utilizando gradientes.

\subsection{Modelos Baseados em \textit{Transformers}}
Os \textit{Transformers} \citep{vaswani2017attention} são modelos de aprendizado profundo que revolucionaram áreas como processamento de linguagem natural (PLN), visão computacional e aprendizado multimodal. Fundamentados no mecanismo de atenção auto-regressiva, esses modelos capturam relações globais entre os elementos de entrada de forma eficiente, superando limitações de arquiteturas tradicionais, como \ac{RNN}s e \ac{LSTM}s, na modelagem de dependências de longo alcance \citep{devlin2018bert}.

Na visão computacional, os \textit{Transformers} foram adaptados por meio dos \ac{ViT}s \citep{dosovitskiy2020vit}. Esses modelos dividem imagens em pequenas janelas chamadas \textit{patches}, que são projetadas em vetores de características antes de serem processadas. Essa abordagem permite a modelagem global das relações espaciais na imagem, reduzindo a dependência de operações convolucionais locais. Com desempenho superior em tarefas como classificação de imagens e detecção de objetos, os \ac{ViT}s destacam-se pela flexibilidade em arquiteturas profundas e pela capacidade de lidar com dados de alta dimensionalidade \citep{touvron2021training}.

No contexto de detecção de \textit{deepfakes}, os \textit{Transformers} têm demonstrado grande eficácia ao explorar inconsistências visuais e temporais em vídeos manipulados \citep{zi2024wilddeepfake}. A capacidade de integrar múltiplas modalidades, como dados visuais e auditivos, os torna adequados para identificar artefatos complexos. Além disso, o mecanismo de atenção permite priorizar regiões de interesse, como olhos e boca, que frequentemente apresentam falhas em \ac{DF}s, aumentando a precisão dos modelos. Com sua versatilidade e desempenho, os \textit{Transformers} representam uma ferramenta essencial para pesquisas e aplicações práticas em detecção de \ac{DF}, acompanhando a evolução das técnicas de manipulação de mídia.

\section{Métricas de Avaliação}
\label{metricas}

As métricas de avaliação são ferramentas essenciais para quantificar a eficácia de um modelo de detecção de \ac{DF}. Elas fornecem uma maneira objetiva de avaliar o desempenho do modelo, permitindo medir sua capacidade de fazer previsões corretas e identificar possíveis erros. Essas métricas ajudam a compreender o quão bem o modelo está realizando sua tarefa e auxiliam na avaliação do seu desempenho \citep{faceli2021}.

\begin{figure}[ht!]
\centering
\includegraphics[width=0.4\textwidth]{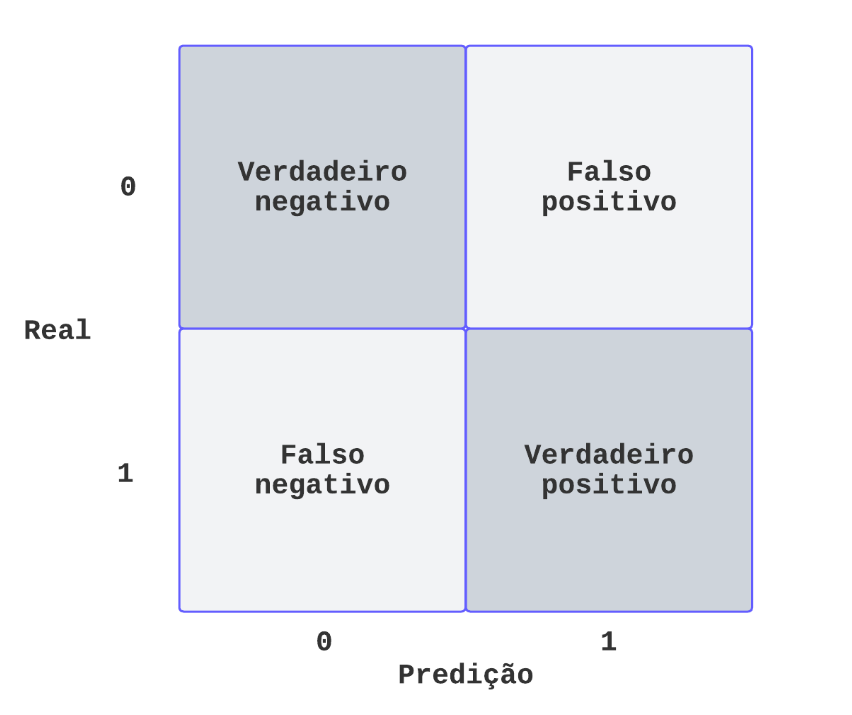}
\caption{Matriz de Confusão.}
\label{fig_matrix_confusao}
\end{figure}

Nesse contexto, a matriz de confusão é uma representação tabular do desempenho de um modelo, exibindo a quantidade de acertos e erros para cada classe. Conforme ilustrado na Figura \ref{fig_matrix_confusao}, a matriz é composta por quatro células principais: Verdadeiros Positivos (VP), Verdadeiros Negativos (VN), Falsos Positivos (FP) e Falsos Negativos (FN) \citep{faceli2021}. A diagonal principal da matriz representa os valores em que o modelo classificou corretamente, enquanto a diagonal secundária mostra os valores em que o modelo fez previsões incorretas. A partir dessa matriz, é possível calcular diversas métricas que ajudam a compreender o desempenho geral do modelo e identificar possíveis áreas de melhoria \citep{Monard2003a}.

As métricas mais utilizadas são descritas a seguir e apresentadas de forma compacta na Tabela \ref{tab:metricas}. A acurácia é a proporção de predições corretas em relação ao total de predições realizadas. A precisão mede a proporção de exemplos classificados corretamente como positivos em relação ao total classificados como positivos, sendo especialmente relevante quando o foco é minimizar falsos positivos. A revocação, também conhecida como sensibilidade ou \textit{recall}, avalia a proporção de exemplos positivos corretamente classificados em relação ao total de exemplos que deveriam ser positivos. O F1-Score combina precisão e revocação em uma única medida, oferecendo uma visão equilibrada do desempenho do modelo. Por fim, a curva \ac{ROC} - Característica de Operação do Receptor, e sua métrica associada, a \ac{AUC} - Área Sob a Curva, fornecem uma representação gráfica da capacidade discriminativa do modelo, indicando o quão bem ele distingue entre classes. Essas métricas são cruciais para avaliar a eficácia de modelos de detecção de \ac{DF}, identificando possíveis limitações e oportunidades de melhoria. O Capítulo \ref{resultados} discute a aplicação prática dessas métricas para detecção e como elas influenciam a análise de desempenho.

\begin{table}[!ht]
\centering
\caption{Métricas de Avaliação}
\label{tab:metricas}
\begingroup
\setlength{\extrarowheight}{6.5pt} 
\resizebox{\textwidth}{!}{
\begin{tabular}{|c|c|m{9cm}|} 
\hline
\rowcolor{lightgray} 
\textbf{Métrica}      & \textbf{Fórmula}                                              & \textbf{Descrição}                                                                                          \\ \hline
Acurácia (\(Acc\))    & \( \text{Acc} = \frac{VP + VN}{VP + VN + FP + FN} \)          & Mede a proporção de predições corretas em relação ao total de amostras avaliadas.                           \\ \hline
Precisão (\(Prec\))   & \( \text{Prec} = \frac{VP}{VP + FP} \)                       & Indica a proporção de exemplos classificados como positivos que realmente são positivos.                    \\ \hline
Revocação (\(Recall\))   & \( \text{Recall} = \frac{VP}{VP + FN} \)                        & Mede a capacidade do modelo de identificar corretamente os exemplos positivos.                              \\ \hline
F1-Score (\(F_1\))    & \( F_1 = 2 \cdot \frac{\text{Precisão} \cdot \text{Revocação}}{\text{Precisão} + \text{Revocação}} \) & Combina precisão e revocação em uma única métrica balanceada.                                               \\ \hline
AUC-ROC               & ---                                                          & Representa a capacidade geral do modelo em distinguir entre classes, medindo a área sob a curva ROC.        \\ \hline
\end{tabular}}
\endgroup
\end{table}

\section{Bases de Dados na Detecção de \textit{Deepfakes}}

As bases de dados desempenham um papel fundamental na evolução da detecção de \ac{DF}, fornecendo cenários diversos e desafiadores para treinar e avaliar modelos de aprendizado de máquina. Conjuntos de dados amplamente utilizados, como \textit{FaceForensics++}, \textit{Celeb-DF}, \textit{DFDC} e \textit{DeepfakeTIMIT}, estabeleceram padrões na área ao introduzir manipulações com diferentes níveis de realismo e técnicas de geração \citep{rossler2019faceforensics, li2020celeb, dolhansky2020deepfake, korshunov2018deepfaketimit}.

Cada conjunto de dados apresenta características distintas que desafiam os modelos de detecção. Por exemplo, o \textit{FaceForensics++} inclui 1.000 vídeos originais manipulados por técnicas como \textit{FaceSwap} \citep{korshunova2017fastfaceswapusingconvolutional}, \textit{Face2Face} \citep{thies2020face2facerealtimefacecapture} e \textit{NeuralTextures} \citep{thies2019deferredneuralrenderingimage}, enquanto o \textit{Celeb-DF} oferece 5.639 vídeos manipulados a partir de clipes de celebridades extraídos do \textit{YouTube}. Já o \ac{DFDC}, com mais de 100.000 vídeos, explora manipulações geradas com técnicas variadas e inclui vídeos com diferentes níveis de compressão. O \textit{DeepfakeTIMIT} é uma base menor, contendo 640 vídeos, focada em manipulações de áudio e vídeo geradas a partir de pares de vídeos sincronizados.

Com o avanço das técnicas de geração de \ac{DF}, surgiram bases como \textit{WildDeepfake} e \textit{DeepSpeak}, que representam o estado da arte em desafios contemporâneos de detecção. Essas bases incluem cenários mais complexos e diversificados, com manipulações avançadas que combinam trocas de rosto e sincronização labial, como no caso do \textit{DeepSpeak}, além de vídeos capturados em ambientes reais, como no \textit{WildDeepfake}. Ao integrar dados com diversidade étnica, de gênero e técnica, essas bases ampliam a robustez dos modelos e avaliam a capacidade de generalização frente a cenários próximos ao mundo real \citep{zi2024wilddeepfake, barrington2024deepspeakdatasetv10}.

A criação de bases de dados para detecção de \ac{DF} geralmente segue um processo que envolve a coleta, manipulação e organização de vídeos ou imagens reais e manipuladas. Inicialmente, vídeos reais são coletados de fontes públicas, como o \textit{YouTube}, ou capturados em ambientes controlados, como no \ac{DFDC}, em que atores contratados participam de gravações. Técnicas de manipulação, como \textit{FaceSwap}, e métodos baseados em \ac{GAN}s e \ac{VAE}s, são aplicadas para criar amostras manipuladas. Esses vídeos passam por etapas adicionais de ajuste de qualidade e compressão para simular cenários reais e desafiadores.

Bases como o \textit{FaceForensics++} utilizam compressão variável (HQ e LQ), enquanto outras, como o \textit{WildDeepfake}, apresentam manipulações criadas a partir de vídeos do mundo real com múltiplas pessoas, expressões faciais variadas e técnicas desconhecidas. Esse nível de variação contribui para a avaliação mais rigorosa dos detectores, testando sua eficácia em identificar padrões visuais e temporais associados a \ac{DF}. Bases recentes, como o \textit{DeepSpeak}, incorporam manipulações multimodais, incluindo inconsistências entre áudio e vídeo, ampliando o escopo das técnicas avaliadas. Ademais, as bases de dados também são organizadas com metadados, incluindo informações sobre identidade, técnica de manipulação e qualidade do vídeo, e divididas em conjuntos de treinamento, validação e teste. Essa organização promove a generalização dos modelos e garante que os detectores sejam avaliados de maneira consistente, reproduzindo cenários que refletem os desafios para detecção.

\section{Trabalhos Relacionados}

A detecção de \textit{deepfakes} envolve uma ampla gama de técnicas que exploram desde características físicas, como bordas de fusão e inconsistências no ambiente, até a identificação de artefatos e anomalias gerados durante o processo de manipulação. Abordagens baseadas em aprendizado de máquina, como \ac{CNN}s, \ac{RNN}s e \ac{VAE}s, têm se destacado na literatura, assim como métodos que utilizam sinais biológicos, propriedades de frequência e inconsistências espaciais e temporais. Apesar dos avanços, o desenvolvimento contínuo de métodos mais sofisticados é essencial para acompanhar a evolução das técnicas de geração de \ac{DF}.

\subsection{Métodos de Detecção de \textit{Deepfake}}
\label{sec:metodos}
Diversos métodos de detecção têm sido propostos na literatura para enfrentar os desafios impostos pelos avanços das técnicas de geração de \ac{DF}. Esses métodos exploram diferentes abordagens, incluindo características visuais, sinais biológicos e modelos baseados em \textit{Transformers}, compondo um panorama abrangente de estratégias.

Entre os métodos visuais, o \textit{MesoNet} \citep{Afchar_2018} utiliza propriedades mesoscópicas para identificar \ac{DF} gerados por técnicas como \textit{FaceSwap} e \textit{Face2Face}. \cite{nguyen2018capsuleforensicsusingcapsulenetworks} propuseram um modelo que combina \textit{VGG-19} com \textit{capsule networks} para detectar falsificações como \textit{FaceSwap} e \textit{Facial Reenactment}, enquanto \cite{yang2018exposingdeepfakesusing} introduziram uma abordagem que compara poses 3D estimadas de marcos faciais para revelar inconsistências em regiões manipuladas.

A \textit{XceptionNet} tem se mostrado eficaz ao explorar convoluções separáveis em profundidade para capturar padrões sutis em dados manipulados, sendo amplamente aplicada em conjuntos como o \textit{FaceForensics++}. Métodos como o \textit{Face X-ray} \citep{li2020facexraygeneralface} detectam bordas de fusão para identificar manipulações, enquanto \cite{sun2022faceforgerydetectionbased} propuseram uma abordagem baseada em grafos espaciais e temporais para capturar trajetórias faciais adulteradas. Adicionalmente, \cite{kim2021generalizedfacialmanipulationdetection} apresentaram um \textit{framework} forense que combina características de cor em nível de pixel e um modelo 3D-\ac{CNN} para análise robusta e generalizada.

Sinais biológicos também têm sido explorados, como o trabalho de \cite{li2018ictuoculiexposingai}, que utiliza redes recursivas e \ac{CNN}s para detectar anomalias em padrões de piscadas de olhos. Já \cite{9304936} modificaram a \textit{XceptionNet}, integrando mapas de fluxo óptico denso para isolar \ac{DF} em níveis de instância e vídeo.

Modelos baseados em \textit{Transformers}, como o \textit{Convolutional Vision Transformer} (CViT) \citep{zhu2022shotfaceswappingmegapixels}, também têm ganhado destaque por explorar características espaciais e temporais. \cite{heo2022} propuseram um \textit{Vision Transformer} aprimorado com vetores concatenados de características extraídas por \ac{CNN}s, enquanto outros estudos combinam \textit{EfficientNet} com \textit{ViTs} \citep{Coccomini_2022}, reforçando a eficácia dessa abordagem para detecção de \ac{DF}.

\subsection{Classificação dos Métodos de Detecção}

De acordo com \cite{mirsky2021creation}, os métodos de detecção de \ac{DF} podem ser classificados em duas categorias principais: aqueles baseados na identificação de artefatos específicos e os que utilizam abordagens indiretas, empregando redes neurais para inferir características relevantes.

\subsubsection{Detecção Baseada em Artefatos}

Os métodos baseados em artefatos exploram inconsistências geradas durante o processo de manipulação de \ac{DF}, dividindo-se em diferentes subcategorias:

\begin{itemize} 
    \item \textbf{Fusão (Espacial):} Artefatos introduzidos na etapa de fusão da face manipulada com o fundo da imagem original. Técnicas comuns incluem detecção de bordas, avaliação de medidas de qualidade e análise no domínio da frequência.
    
    \item \textbf{Ambiente (Espacial):} Inconsistências entre o conteúdo gerado e o original, como iluminação, transformações geométricas e fidelidade. Métodos nessa categoria utilizam \ac{CNN}s para comparar regiões da face e do fundo ou empregam redes \ac{ED}s para gerar representações das partes e do contexto, alimentando classificadores com essas diferenças.
    
    \item \textbf{Análise Forense (Espacial):} Padrões sutis, como artefatos específicos gerados por redes \ac{GAN}, podem ser detectados mesmo em presença de compressão ou ruído. Abordagens incluem análise de resíduos com \ac{ED}s aprimoradas por filtros, integradas a redes \ac{LSTM} para classificar sequências de quadros. Alternativamente, algumas técnicas enfatizam ruídos e imperfeições como pré-processamento para classificadores.
    
    \item \textbf{Comportamento (Temporal):} Padrões comportamentais ou anomalias são explorados para detectar manipulações, como discrepâncias entre emoções extraídas de áudio e vídeo. Redes Siamesas, projetadas para medir a similaridade entre pares de dados, também são utilizadas para avaliar a consistência temporal e identificar possíveis manipulações.
    
    \item \textbf{Fisiologia (Temporal):} A ausência de sinais fisiológicos, como frequência cardíaca, pulso ou padrões naturais de piscadas, é utilizada como indicador de falsificações. Técnicas baseadas em monitoramento de sinais fisiológicos têm se mostrado promissoras para a detecção de \ac{DF}.
    
    \item \textbf{Sincronização (Temporal):} Inconsistências em ataques de sincronização labial, como correlações incorretas entre fonemas e visemas, são evidenciadas ao analisar os movimentos da boca em relação ao áudio.
    
    \item \textbf{Coerência (Temporal):} Artefatos de incoerência temporal, como efeitos de \textit{flicker} ou \textit{jitter} na região facial, são detectados por redes \ac{RNN} ou \ac{LSTM}, que avaliam fluxos ópticos ou erros de reconstrução em quadros sucessivos.
\end{itemize}

\subsubsection{Detecção Baseada em Abordagens Indiretas}

Os métodos baseados em abordagens indiretas utilizam redes neurais para extrair características relevantes e podem ser classificados em duas categorias principais:
\begin{itemize} 
    \item \textbf{Classificação:} Redes \ac{CNN}s, siamesas, modelos hierárquicos com \ac{GRU} bidirecional e mecanismos de atenção, \textit{ensembles} de classificadores e \ac{CNN}s 3D são amplamente empregados para detectar \ac{DF} em imagens e vídeos.    
    \item \textbf{Detecção de Anomalias:} Técnicas que tratam \ac{DF} como anomalias incluem redes treinadas exclusivamente com dados reais, medindo ativações de redes de reconhecimento facial, analisando erros de reconstrução em \ac{VAE}s ou comparando imagens com suas projeções no espaço latente.
\end{itemize}

\section{Estado da Arte}

A detecção de \textit{Deepfakes} é um campo em constante desenvolvimento, impulsionado por desafios decorrentes do avanço das técnicas de geração de vídeos falsos. Dentro desse contexto, várias arquiteturas têm sido propostas, cada uma com diferentes abordagens e focos. Este trabalho enfatiza a análise do modelo \textit{\ac{GenConViT}}, além de discutir os principais modelos incluídos no \textit{DeepfakeBenchmark}, uma ferramenta utilizada para avaliação de detectores de \ac{DF}.

\subsection{\textit{GenConViT} para Detecção de Deepfakes}

O modelo \textit{\ac{GenConViT}} combina arquiteturas modernas, como \ac{AE}, \ac{VAE}, \textit{ConvNeXt} \citep{liu2022convnet2020s} e \textit{Swin Transformer} \citep{liu2021swintransformerhierarchicalvision}, para superar limitações na generalização de detectores de \ac{DF} \citep{wodajo2023deepfake}. O modelo em questão é estruturado em duas redes independentes: a Rede A, que utiliza \ac{AE} para reconstrução e \textit{ConvNeXt} e \textit{Swin Transformer} para extração de características e classificação binária com a função de perda de \ac{CE}; e a Rede B, que emprega \ac{VAE} para modelagem probabilística no espaço latente, além de \textit{ConvNeXt e Swin Transformer}, utilizando uma combinação de \ac{CE} e \ac{MSE} para ajuste de reconstruções e predições. Ambas as redes processam imagens pré-dimensionadas de \(224 \times 224 \times 3\), e suas predições são combinadas durante a inferência. O modelo demonstrou alta eficácia em bases como DFDC, FaceForensics++ e Celeb-DF v2, destacando-se por sua robustez e capacidade de generalização.

\subsection{\textit{DeepfakeBenchmark}}
O \textit{DeepfakeBenchmark} é uma plataforma que visa padronizar a avaliação de detectores de \ac{DF}. Ele integra um conjunto diversificado de 15 modelos de detecção, incluindo arquiteturas clássicas como \textit{XceptionNet}, \textit{EfficientNet}, e \textit{Meso4Inception}, além de novas arquiteturas como \textit{SPSL} e \textit{UCF}. 

Em síntese, o \textit{XceptionNet}, conforme a Seção \ref{sec:metodos}, utiliza convoluções separáveis em profundidade para capturar padrões sutis em imagens, sendo especialmente eficaz na identificação de artefatos gerados. O \textit{EfficientNet}, por sua vez, aplica uma abordagem de escalonamento composto para ajustar a largura, profundidade e resolução da rede, permitindo detectar manipulações. Já o \textit{Meso4Inception} representa uma das implementações da \textit{MesoNet} e concentra-se em padrões mesoscópicos, explorando características de média escala, como texturas e bordas, para identificar inconsistências em regiões manipuladas. Ademais, o Spatial-Phase Shallow Learning (SPSL) detecta \ac{DF} analisando artefatos de manipulação facial por meio da fase espectral, complementada com representações RGB. Ele utiliza redes rasas para focar em texturas locais, promovendo maior generalização a novas técnicas de manipulação. O modelo \textit{Uncovering Common Features} (UCF) detecta \ac{DF} identificando características comuns compartilhadas entre diferentes métodos de falsificação. Utilizando um framework multitarefa, ele separa padrões específicos de falsificação, características irrelevantes e traços gerais de \ac{DF}, garantindo maior capacidade de generalização. Técnicas como \textit{Adaptive Instance Normalization} (AdaIN) e perdas contrastivas reforçam a separação das características, enquanto perdas de reconstrução e classificação aumentam a robustez do modelo.

Além disso, o \textit{benchmark} fornece um ambiente de teste unificado, abordando desde o pré-processamento dos dados até a análise das métricas de desempenho. A ferramenta promove a reprodutibilidade e comparabilidade entre diferentes abordagens, permitindo a exploração de novas estratégias na detecção de \textit{Deepfakes}. Cada modelo no \textit{DeepfakeBenchmark} é avaliado com base em métricas padronizadas, como \ac{AUC}, precisão e \textit{recall}.

\chapter{Desenvolvimento}
\label{desenvolvimento}

\begin{figure}[!ht]
\centering
\includegraphics[width=1\textwidth]{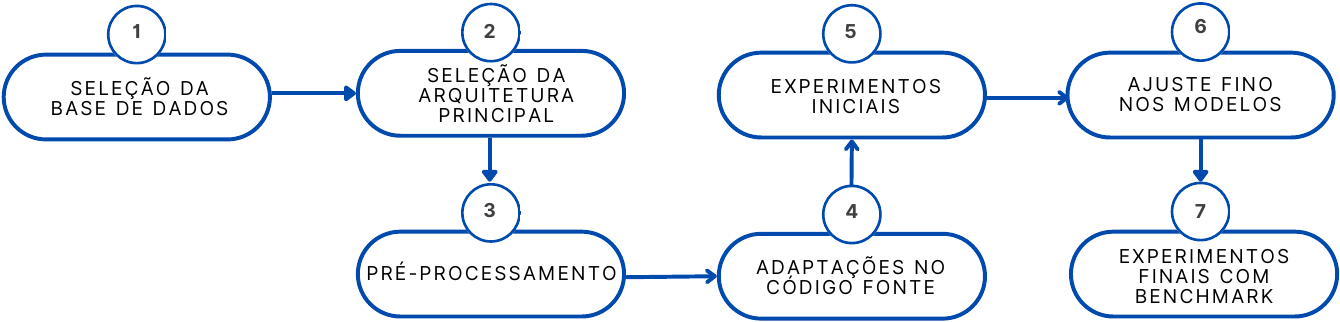}
\caption{Metodologia}
\label{fig_metodologia}
\end{figure}

O desenvolvimento deste trabalho seguiu uma série de etapas estruturadas com o objetivo de comparar o estado da arte dos modelos de detecção de \ac{DF} sem e com ajuste fino. As etapas incluíram a seleção da base de dados, seleção da arquitetura principal, adaptações no código fonte, pré-processamento, experimentos iniciais, ajuste fino nos modelos e, por fim, os experimentos finais com \textit{benchmark}, conforme o fluxo presente na Figura \ref{fig_metodologia}. As implementações necessárias para cada etapa utilizaram a linguagem \textit{Python}. Nesse sentido, toda a estrutura do projeto foi disponibilizada em um repositório remoto
\footnote{\href{https://github.com/MatMB115/tfg-deepfake-detection}{https://github.com/MatMB115/tfg-deepfake-detection}}, facilitando a reprodutibilidade e o acesso aos códigos utilizados.

\section{Base de Dados}
\label{sec:base_dados}
Neste trabalho, foram selecionadas duas bases de dados principais: \textit{WildDeepfake} \citep{zi2024wilddeepfake} e \textit{DeepSpeak} \citep{barrington2024deepspeakdatasetv10}. A escolha da \textit{WildDeepfake} se deu pela natureza variada dos dados, que inclui vídeos capturados em cenários reais, apresentando um desafio significativo para os modelos de detecção de \ac{DF}. Já o \textit{DeepSpeak} foi escolhido devido à sua preocupação em abranger diferentes métodos de geração de \ac{DF}, além de contemplar a diversidade étnica nos dados, garantindo uma análise mais inclusiva e realista. Embora ambas as bases sejam recentes e relevantes para o campo, apenas a \textit{DeepSpeak} foi utilizada no fluxo de comparação entre \textit{\ac{GenConViT}} e os modelos do \textit{benchmark} ao decorrer dos experimentos.

Diferente dos conjuntos presentes na literatura, como \textit{FaceForensics++} \citep{rossler2019faceforensics}, \textit{Celeb-DF} (V1 e V2) \citep{li2020celeb}, \ac{DFDC} \citep{dolhansky2020deepfake}, e \textit{DeepfakeTIMIT} \citep{korshunov2018deepfaketimit}, conforme a Tabela \ref{tab:datasets_comparison}, que são frequentemente criados em ambientes controlados e onde há modelos que já apresentam métricas excepcionais, essas novas bases de dados, como a \textit{WildDeepfake} e a \textit{DeepSpeak}, representam o estado atual da geração de \ac{DF}. Elas proporcionam cenários mais desafiadores e diversificados, levando a modelos de detecção mais robustos e capazes de generalizar melhor para novas técnicas de forjamento que ainda não foram amplamente exploradas.

O \textit{WildDeepfake} é composto por 7.314 sequências de faces extraídas de 707 vídeos \ac{DF}, todos coletados da internet. Este conjunto se destaca por incluir \ac{DF} do mundo real, criados em uma ampla variedade de cenários, com múltiplas pessoas e expressões faciais, e com técnicas desconhecidas. Ademais, utiliza vídeos com alta qualidade visual e maior realismo, o que torna mais difícil para os detectores treinados em outros conjuntos generalizarem para ele.

\begin{table}[ht!]
\centering
\caption{Comparação de bases de dados para detecção de \textit{Deepfake} com referências.}
\resizebox{\textwidth}{!}{%
\begin{tabular}{|l|c|c|c|c|c|c|c|c|}
\hline
\rowcolor{lightgray} 
\textbf{Nome} & \textbf{Identidades Únicas} & \textbf{Filmagens Originais} & \textbf{Consentimento} & \textbf{\textit{Faceswap}} & \textbf{\textit{Lipsync}} & \textbf{Áudio} & \textbf{Filmagens Deepfake} & \textbf{Referência} \\
\hline
Deepfake-TIMIT & 32 & 320 & Sim & \ding{51} & - & - & LQ:320 HQ:320 & \citep{korshunov2018deepfaketimit}  \\
FaceForensics & ? & 1,004 & Não & \ding{51} & - & - & 2,008 & \citep{rossler2018faceforensics} \\
FaceForensics++ & 997 & 1,000 & Não & \ding{51} & \ding{51} & - & 4,000 & \citep{rossler2019faceforensics} \\
DFDC & 3,426 & 23,654 & Sim & \ding{51} & - & - & 104,500 & \citep{dolhansky2020deepfake} \\
Celeb-DF & 59 & 590 & Não & \ding{51} & - & - & 5,639 & \citep{li2020celeb} \\
DeeperForensics & 100 & 50,000 & Sim & \ding{51} & - & - & 10,000 & \citep{jiang2020deeperforensics} \\
FakeAVCeleb & 600 & 570 & Não & \ding{51} & \ding{51} & - & 25,000 & \citep{khalid2022fakeavceleb} \\
ForgeryNet & 5,400 & 99,630 & Sim/Não & \ding{51} & - & - & 121,617 & \citep{he2021forgerynet} \\
LAV-DF & 153 & 36,431 & Não & - & \ding{51} & - & 99,873 & \citep{cai2023reallymeanthatcontent} \\
\textbf{WildDeepfake} & 707 & 3,805 & Não & ? & ? & ? & 3,509 & \citep{zi2024wilddeepfake} \\
\textbf{DeepSpeak v1.0} & 220 & 6,226 & Sim & \ding{51} & \ding{51} & \ding{51} & 6,799 & \citep{barrington2024deepspeakdatasetv10} \\
\hline
\end{tabular}%
}
\label{tab:datasets_comparison}
\end{table}

\begin{figure}[ht!]
    \centering
    \includegraphics[width=0.9\textwidth]{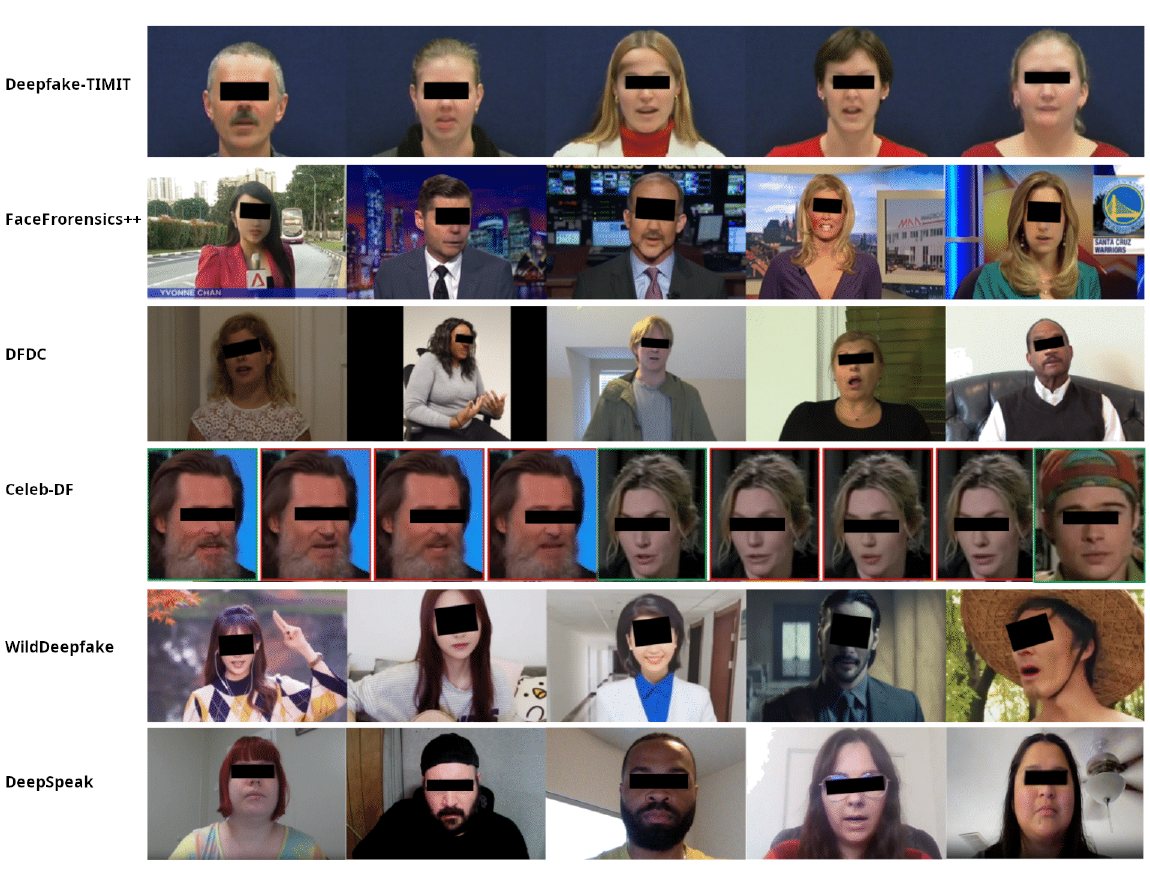}
    \caption{\textit{WildDeepfake} e \textit{DeepSpeak} comparados com quatro outros conjuntos de dados existentes. As cenas em \textit{WildDeepfake} são mais diversificadas e os rostos falsos parecem mais realistas, refletindo o cenário desafiador do mundo real. O \textit{DeepSpeak}, por sua vez, foca em DF de sincronização labial, trazendo dados recentes de fala e gestos. Para proteger a privacidade, a região dos olhos foi coberta nas imagens.}
    \label{fig:datasets_comparison}
\end{figure}

Já o \textit{DeepSpeak} é um conjunto de dados que visa apoiar a pesquisa forense digital, fornecendo um conjunto abrangente de áudio e vídeo de pessoas falando e gesticulando em frente a câmeras. Ele contém 44 horas de gravações, sendo 17 horas de vídeos reais capturados de 220 indivíduos diversos, e 26 horas de \ac{DF} gerados utilizando técnicas de \textit{face-swap} como \textit{FaceFusion} \citep{ruhs2024facefusion}, \textit{FaceFusion + GAN} \citep{zhou2022codebook} e \textit{FaceFusion Live}, além de técnicas de \textit{lip-sync} como \textit{Wav2Lip} \citep{prajwal2020wav2lip} e \textit{VideoRetalking} \citep{cheng2022videoretalking}. Além disso, o conjunto é organizado em trechos para treinamento e teste, e inclui \ac{DF} com vozes naturais e geradas por IA, ampliando a complexidade dos cenários. O \textit{DeepSpeak} se destaca por sua diversidade étnica e de gênero, oferecendo um desafio robusto e inclusivo para a detecção abordando tanto aspectos visuais quanto inconsistências auditivas. Ao abranger uma variedade de métodos modernos de geração de \ac{DF}, esse conjunto proporciona uma oportunidade única para avaliar o desempenho de modelos em cenários mais desafiadores e realistas. A Figura \ref{fig:datasets_comparison} apresenta uma comparação visual entre os principais conjuntos, incluindo o \textit{WildDeepfake} e o \textit{DeepSpeak}.

\section{Arquitetura Principal: \textit{GenConViT}}

\begin{figure}[ht!]
    \centering
    \includegraphics[width=0.72\textwidth]{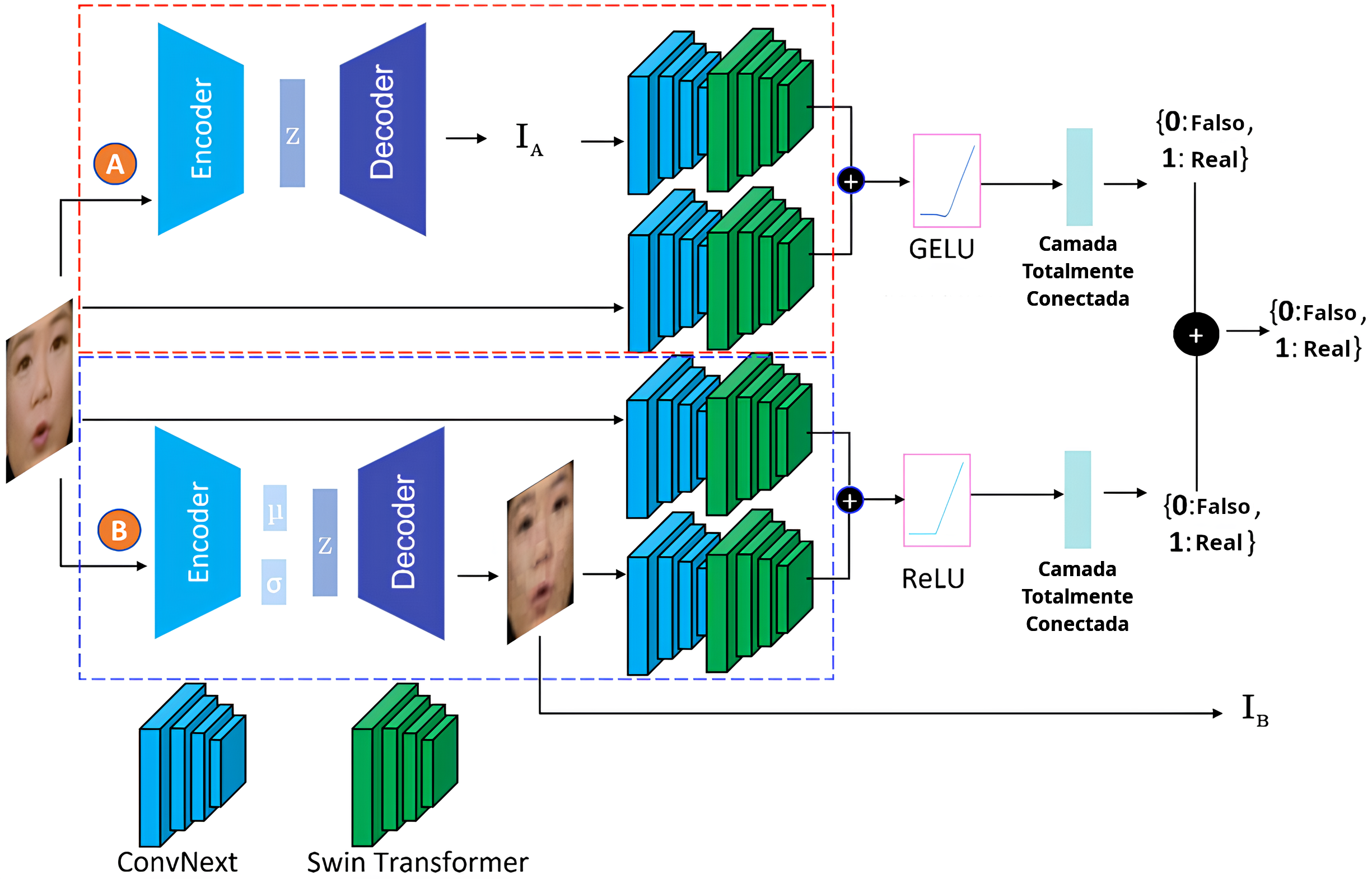}
    \caption{Arquitetura do \textit{GenConViT}. Fonte: adaptado de \cite{wodajo2023deepfake}}
    \label{fig:genconvit_arch}
\end{figure}

O \textit{\ac{GenConViT}} é o modelo principal estudado e adaptado neste trabalho, composto por duas redes distintas, chamadas de \textbf{Rede A} e \textbf{Rede B}, como ilustrado na Figura~\ref{fig:genconvit_arch}. Ambas integram \ac{CNN}s e \textit{Transformers}, utilizando o \textit{ConvNeXt} \citep{liu2022convnet2020s} e o \textit{Swin Transformer} \citep{liu2021swintransformerhierarchicalvision} como \textit{backbones}. Essa combinação permite a extração de características visuais detalhadas e o processamento de representações locais e globais, fundamentais para a detecção de \ac{DF}.

A \textbf{Rede A} utiliza um \ac{AE} que comprime imagens de entrada em um espaço latente \(z\) e as reconstrói posteriormente. O \textit{encoder} do \ac{AE} é composto por cinco camadas convolucionais com funções de ativação \ac{ReLU} e gera um espaço latente de tamanho \(256 \times 7 \times 7\). O \textit{decoder}, por sua vez, consiste em cinco camadas de convoluções transpostas, reconstruindo as imagens no formato \(224 \times 224 \times 3\). Essa estrutura é projetada para capturar padrões visuais sutis presentes em manipulações.

Já a \textbf{Rede B} utiliza um \ac{VAE}, que se diferencia por modelar o espaço latente como uma distribuição probabilística, permitindo capturar incertezas nas representações. O \textit{encoder} do \ac{VAE} é composto por quatro camadas convolucionais com normalização por lotes e função de ativação \textit{LeakyReLU}, produzindo uma representação latente com tamanho \(12544\). O \textit{decoder} é semelhante ao da Rede A, mas reconstrói as imagens em uma resolução reduzida de \(112 \times 112 \times 3\). Essa capacidade de modelagem probabilística torna a Rede B mais eficaz para lidar com vídeos que apresentam maior variação temporal.

Ambas as redes utilizam uma arquitetura híbrida \textit{ConvNeXt-Swin Transformer}. O \textit{ConvNeXt} extrai características visuais detalhadas tanto das imagens originais quanto das reconstruções geradas pelo \ac{AE}/\ac{VAE}, enquanto o \textit{Swin Transformer} processa essas características por meio de atenção local e global. O módulo \textit{HybridEmbed} reduz a dimensionalidade dessas características para 768 por meio de uma convolução \(1 \times 1\). Em seguida, as características são achatadas e transpostas em vetores sequenciais, que alimentam o \textit{Swin Transformer} para processamento avançado. A Rede A processa características das imagens originais e reconstruções do \ac{AE}, utilizando a função \textit{GELU} como ativação em suas camadas totalmente conectadas, enquanto a Rede B realiza o mesmo processo com representações probabilísticas do \ac{VAE} e utiliza a função \textit{ReLU}. As saídas de ambas as redes híbridas são concatenadas e passam por uma camada linear para gerar predições binárias (\textit{real} ou \textit{fake}). Além disso, a Rede B também reconstrói as imagens de saída.

O \textit{\ac{GenConViT}} foi projetado para maximizar a capacidade de detecção de \ac{DF} em vídeos, utilizando entradas com resolução \(224 \times 224\) \textit{pixels}. A combinação de imagens brutas e representações latentes possibilita a captura de inconsistências visuais e padrões ocultos no espaço latente. No trabalho original, o modelo foi treinado em conjuntos de dados amplamente utilizados, incluindo \textit{DFDC} \citep{dolhansky2020deepfake}, \textit{FaceForensics++} \citep{rossler2019faceforensics}, \textit{Celeb-DF} (v2) \citep{li2020celeb} e \textit{Deepfake-TIMIT} \citep{korshunov2018deepfaketimit}.

Além disso, para avaliar o desempenho do \textit{\ac{GenConViT}} de forma comparativa, foi empregado o \textit{DeepfakeBenchmark} \citep{DeepfakeBench_YAN_NEURIPS2023}. Esta plataforma  permitiu a comparação do \textit{\ac{GenConViT}} com outros modelos de detecção incluindo o \textit{EfficientB4} \citep{tan2019efficientnet}, \textit{Meso4Inception} \citep{Afchar_2018}, \textit{XceptionNet} \citep{rossler2019faceforensics}, \textit{\ac{SPSL}} \citep{liu2021spatial} e \textit{\ac{UCF}} \citep{yan2023ucfuncoveringcommonfeatures}. Vale ressaltar que esses modelos foram selecionados com base nas melhores métricas de desempenho disponibilizadas pelo \textit{Benchmark}. Nesse sentido, as arquiteturas supracitadas foram implementadas, em seus trabalhos de origem, utilizando o \textit{framework PyTorch} \citep{paszke2019pytorchimperativestylehighperformance}. Da mesma forma, foram utilizadas bibliotecas de extração de características de imagens, com ênfase no reconhecimento facial, incluindo \textit{Dlib} \citep{dlib09}, \textit{face\_recognition} \citep{face_recognition}, \textit{Mediapipe} \citep{lugaresi2019mediapipeframeworkbuildingperception} e \textit{OpenCV} \citep{itseez2015opencv}, esta última amplamente empregada para manipulação de imagens.

\section{Pré-processamento}
Os conjuntos de dados utilizados neste trabalho passaram por um pré-processamento cuidadoso para garantir a compatibilidade com o \textit{\ac{GenConViT}} e \textit{DeepfakeBenchmark}. Todos os dados pré-processados de faces extraídas e vídeos serão disponibilizados para download no \textit{Kaggle}, uma plataforma de aprendizado de máquina do \textit{Google}, permitindo a reprodutibilidade dos experimentos. Vale ressaltar que foi realizada a sugestão de \cite{barrington2024deepspeakdatasetv10} para anonimizar os nomes dos arquivos durante o pré-processamento e verificar se essa característica influenciava o desempenho dos modelos. Esse teste foi implementado para garantir que nenhuma informação não relacionada ao conteúdo visual dos vídeos estivesse afetando os resultados. As etapas específicas de pré-processamento para cada conjunto estão resumidas na Tabela \ref{tab:preprocessamento}:

\begin{table}[ht!]
    \centering
    \caption{Pré-processamento dos conjuntos de dados para treinamento no \textit{GenConViT} e \textit{DeepfakeBenchmark}}
    \resizebox{\textwidth}{!}{%
    \begin{tabular}{|l|c|c|c|c|c|c|}
    \hline
    \rowcolor{lightgray}
    \textbf{Base de Dados} & \textbf{Arquitetura} & \textbf{Ferramenta de Extração} & \textbf{Resolução} & \textbf{Quadros por Vídeo} & \textbf{Divisão (Treino/Validação/Teste)} & \textbf{Total de Quadros} \\ \hline
    WildDeepfake & GenConViT & Nativo do Conjunto & \textbf{224x224} & Sequências prontas & \textbf{87\%/10\%/3\%} & 189.713 \\ \hline
    DeepSpeak & GenConViT & MediaPipe & \textbf{224x224} & Até 15 quadros & \textbf{80\%/15\%/5\%} & 150.681 \\ \hline
    DeepSpeak & DeepfakeBenchmark & dlib & \textbf{256x256} & Até 15 quadros & \textbf{80\%/15\%/5\%} & 137.495 \\ \hline
    \end{tabular}%
    }
    \label{tab:preprocessamento}
\end{table}

\subsection{WildDeepfake para o \textit{GenConViT}}
O \textit{WildDeepfake} já fornece as sequências de vídeos com as faces extraídas e redimensionadas no formato de 224x224 \textit{pixels}, utilizando três canais de cor \ac{RGB}. Isso simplificou o pré-processamento, pois as imagens já estavam preparadas para serem usadas diretamente pelo \textit{\ac{GenConViT}}. A principal tarefa realizada foi a reorganização e divisão dos dados. A partir da parcela designada pelo próprio conjunto de dados para testes, foram extraídas aleatoriamente 863 sequências de faces para avaliação. Para o treinamento, foram selecionados 189.713 quadros do conjunto separado para esse fim. Esses quadros foram organizados nas proporções de 87\% para treinamento, 10\% para validação, e 3\% para testes, garantindo uma distribuição semelhante ao que foi proposto para treinamento no trabalho do \textit{\ac{GenConViT}}.

\subsection{DeepSpeak para o \textit{GenConViT}}
\label{sec:deepspeak_genconvit}
Para o \textit{DeepSpeak}, foi necessário um pré-processamento mais elaborado, que incluiu a extração de faces diretamente dos vídeos. Foram selecionados 1.472 vídeos (sequências) do conjunto de dados para testes fornecidos. Utilizando bibliotecas como \textit{OpenCV} \citep{opencv_library} e \textit{MediaPipe} \citep{lugaresi2019mediapipeframeworkbuildingperception}, as faces foram detectadas, extraídas e redimensionadas para o formato de 224x224 \textit{pixels}, também utilizando três canais de cor \ac{RGB}, para manter a compatibilidade com o \textit{\ac{GenConViT}}. Dos 10.202 vídeos fornecidos para treinamento (sendo 5.300 vídeos falsos e 4.902 vídeos reais), foram extraídos até 15 quadros com faces detectadas de cada vídeo, assegurando a representatividade das manipulações. No total, foram extraídos 150.681 quadros aleatoriamente, que foram divididos nas proporções de 80\% para treinamento, 15\% para validação, e 5\% para testes. Essa abordagem garantiu uma ampla diversidade nos dados processados, o que é essencial para melhorar o desempenho do \textit{\ac{GenConViT}}.

\subsection{DeepSpeak para os Modelos do DeepfakeBenchmark}
Para o \textit{DeepfakeBenchmark}, foram selecionados os mesmo 1.472 vídeos do conjunto para testes da Seção \ref{sec:deepspeak_genconvit}. o pré-processamento das faces foi realizado utilizando as ferramentas internas do \textit{benchmark}, que empregam a biblioteca \textit{dlib} \citep{dlib09} para a extração das faces. O método usado foi o \textit{shape\_predictor\_81\_face\_landmarks}, responsável pela detecção precisa dos pontos faciais. Para cada vídeo em que uma face foi detectada, foram extraídos até 15 quadros, com um total de 137.495 quadros extraídos. As faces foram redimensionadas para o formato de 256x256 \textit{pixels}, mantendo uma resolução adequada para os modelos do \textit{benchmark}. Após a extração, os quadros foram divididos na proporção de 80\% para treinamento, 15\% para validação, e 5\% para testes.

\section{Ajustes nos Códigos Fonte}
As adaptações realizadas no código do \textit{\ac{GenConViT}} e do \textit{DeepfakeBenchmark} compartilham várias etapas em comum. Ambos os códigos foram modificados para permitir o armazenamento dos resultados das predições, garantindo que os dados pudessem ser reutilizados para análises detalhadas. Nessa perspectiva, os dois códigos exigiram a criação de pipelines específicos para manipular o \textit{DeepSpeak}, com ajustes no processo de extração de faces e no controle sobre o número de quadros utilizados para predição. As alterações foram feitas em repositório ramificados (\textit{fork}) dos trabalhos originais.

\section{Experimentos Iniciais}
\label{sec:exp_preliminar}
Foram realizadas predições utilizando os melhores pesos fornecidos para as redes do \textit{\ac{GenConViT}} e para os modelos do \textit{DeepfakeBenchmark}, aplicados ao \textit{DeepSpeak}, utilizando os 1.472 vídeos previamente selecionados, com 15 quadros por vídeo. É importante destacar que esses pesos foram resultantes do treinamento com os conjuntos de dados da literatura mencionados na Seção \ref{sec:base_dados}. As predições foram conduzidas em um sistema \textit{EndeavourOS} com processador - \ac{CPU} \textit{Intel Xeon E5-2650 v3} e 32GB de \ac{RAM}. Vale destacar que o conjunto de dados \textit{WildDeepfake} foi testado apenas com o \textit{\ac{GenConViT}}, devido à natureza das suas amostras, que são exclusivamente imagens, o que se ajusta ao pipeline de predição direto com imagens que foi implementado no modelo.

Ademais, foram conduzidos experimentos específicos com o \textit{\ac{GenConViT}}, nos quais variou-se a quantidade de quadros usados para predição entre 10, 15 e 24 quadros por vídeo, a fim de avaliar o impacto desse parâmetro no desempenho do modelo. Também foi medido o tempo médio de predição das redes \ac{VAE} e \ac{AE}, fornecendo informações sobre a eficiência computacional de cada abordagem. Outro aspecto investigado foi quais métodos presentes no \textit{DeepSpeak} foram mais eficazes em enganar os modelos sem ajustes finos. Essas análises permitiram identificar quais técnicas de geração de \ac{DF} representam os maiores desafios para os modelos em seus estados iniciais, sem otimizações específicas.

As principais métricas utilizadas para a avaliação dos modelos durante os experimentos iniciais foram \textbf{Acurácia}, \textbf{\textit{\ac{AUC}} - \ac{ROC}} com limiar de decisão de 0,5, \textit{F1-Score}, Precisão e \textit{Recall}. Para calcular essas métricas, foi utilizado o módulo \textit{metrics} da biblioteca \textit{scikit-learn} \citep{scikit-learn}, que oferece implementações eficientes e confiáveis dessas funções de avaliação. A utilização do \textit{sklearn} garantiu a padronização no cálculo das métricas e facilitou a comparação entre os diferentes modelos e configurações testadas.

\section{Detalhes de Implementação do Ajuste Fino}

O ajuste fino foi realizado carregando os \textit{checkpoints} - pesos disponibilizados por cada modelo pré-treinado com o objetivo de adaptar os modelos do \textit{GenConViT} e do \textit{DeepfakeBenchmark} ao conjunto de dados \textit{DeepSpeak}. Os detalhes para cada cenário são descritos nas seções \ref{sec:ajustegenconvit} e \ref{sec:ajustebenchmark}.

\subsection{Detalhes do \textit{GenConViT}}
\label{sec:ajustegenconvit}
O modelo \textit{\ac{GenConViT}} é baseado em uma arquitetura híbrida que combina redes convolucionais com \textit{transformers}, utilizando os modelos \textit{ConvNeXt} e \textit{Swin Transformer} como \textit{backbones}. Ambos foram pré-treinados no \textit{ImageNet-1k} \citep{5206848} e a versão utilizada durante o treinamento foi a \textit{tiny}. Esses modelos foram carregados usando a biblioteca \textit{timm} \citep{rw2019timm}, que facilita o acesso a arquiteturas e a seus pesos pré-treinados. O uso desses \textit{backbones} pré-treinados foi fundamental para reduzir o tempo de treinamento, permitindo que o modelo \textit{\ac{GenConViT}} iniciasse o ajuste fino com representações de características já aprendidas, acelerando o processo de convergência para os conjuntos \textit{WildDeepfake} e \textit{DeepSpeak}.

O ambiente de treinamento do \textit{\ac{GenConViT}} foi configurado como uma instância de notebook do \textit{Kaggle}, equipada com duas \ac{vCPU}, uma \ac{GPU} P100 e 15 GB de memória \ac{RAM}. O uso de uma \ac{GPU} permite aproveitar o \textit{\ac{CUDA}}, uma plataforma de computação paralela desenvolvida pela NVIDIA, que acelera significativamente o processamento de tarefas computacionalmente intensivas, como o treinamento de redes neurais, ao distribuir as operações entre os núcleos da \ac{GPU}.

Durante o treinamento, foram aplicadas diversas técnicas de aumentação de dados com uma taxa de aumentação de 90\% para melhorar a robustez do modelo, garantir a diversidade do conjunto de dados de entrada e prevenir \textit{overfitting}. As técnicas de aumentação foram implementadas utilizando a biblioteca \textit{Albumentations} \citep{info11020125} e incluem as seguintes transformações:

\begin{multicols}{2}
\small
\begin{itemize}
    \item \textbf{RandomRotate}: Rotação aleatória das imagens.
    \item \textbf{Transpose}: Transposição dos eixos das imagens.
    \item \textbf{HorizontalFlip}: Espelhamento horizontal.
    \item \textbf{VerticalFlip}: Espelhamento vertical.
    \item \textbf{GaussNoise}: Adição de ruído gaussiano nas imagens.
    \item \textbf{ShiftScaleRotate}: Translação, escala e rotação combinadas.
    \item \textbf{CLAHE}: Equalização adaptativa de histograma.
    \item \textbf{Sharpen}: Aumento da nitidez das imagens.
    \item \textbf{IAAEmboss}: Aplicação de relevo nas imagens.
    \item \textbf{RandomBrightnessContrast}: Ajuste de brilho e contraste.
    \item \textbf{HueSaturationValue}: Modificação de matiz, saturação e valor.
\end{itemize}
\end{multicols}

O ajuste fino das Redes A e B foi realizado utilizando configurações semelhantes em diversos aspectos, as redes empregam uma taxa de aprendizado de 0,0001 e o otimizador \textit{Adam}. Além disso, a função de perda \ac{CE} foi aplicada para ambas, uma vez que realizam a tarefa de classificação binária. No entanto, houve diferenças nas configurações específicas de treinamento. A Rede A - \ac{AE} foi treinada por 4, 5, 8 e 10 épocas, com um tamanho de lote de 32, enquanto a Rede B foi treinada por 4, 5, 6, 8 e 10 épocas, com um tamanho de lote menor de 16. Além da \ac{CE}, a Rede B - \ac{VAE} também utilizou a função de perda \ac{MSE} para minimizar a diferença entre as imagens originais e as reconstruídas, devido à sua capacidade de realizar tanto a classificação quanto a reconstrução das imagens. A Tabela \ref{tab:comparison_A_B} sintetiza as informações apresentadas.

\begin{table}[ht!]
\centering
\caption{Detalhes do Treinamento das Redes A e B do \textit{GenConViT}}
\scriptsize
\begin{tabular}{|c|c|c|}
\hline
\rowcolor{lightgray}
\textbf{Característica}     & \textbf{Rede A}  & \textbf{Rede B} \\ \hline
\textbf{Arquitetura}        & Autoencoder (AE)     & Variational Autoencoder (VAE) \\ \hline
\textbf{Funções de Perda}   & Cross Entropy        & Cross Entropy + MSE \\ \hline
\textbf{Tamanho do Lote}    & 32                  & 16 \\ \hline
\textbf{Épocas}             & 4, 5, 8 e 10               & 4, 5, 6, 8 e 10 \\ \hline
\textbf{Tarefa}             & Classificação       & Classificação + Reconstrução \\ \hline
\textbf{Otimizador}         & Adam                & Adam \\ \hline
\textbf{Taxa de Aprendizado} & 0,0001              & 0,0001 \\ \hline
\textbf{\textit{Backbones}}          & ConvNeXt + Swin Transformer & ConvNeXt + Swin Transformer  \\ \hline
\textbf{Ambiente}           & Kaggle & Kaggle \\ \hline
\end{tabular}%
\label{tab:comparison_A_B}
\end{table}

\subsection{Detalhes dos Modelos do DeepfakeBenchmark}
\label{sec:ajustebenchmark}
Todos os modelos utilizados no \textit{DeepfakeBenchmark} foram treinados e testados com o conjunto de dados \textit{DeepSpeak}. As imagens estão na resolução de 256x256 \textit{pixels}, e cada vídeo foi processado com 15 quadros. O treinamento foi realizado com um tamanho de lote variando entre 16 e 32 para os diferentes modelos. A maioria dos modelos utilizou técnicas de aumentação de dados, incluindo \textit{flip}, \textit{blur}, ajuste de brilho e contraste, além de compressão de qualidade, com exceção do \textit{Meso4Inception}. O otimizador utilizado foi o \textit{Adam}, com uma taxa de aprendizado de 0,0002 e a função de perda \ac{CE}. O ajuste foi conduzido por 5 épocas, conforme a Tabela \ref{tab:comparison_model_DFB}. O treinamento foi realizado no \textit{Google Cloud Computing}, especificamente utilizando o serviço \textit{Vertex AI Workbench}. Neste ambiente, foi configurada uma máquina do tipo \textit{n1-standard-8}, composta por 8 \ac{vCPU}, 30GB de \ac{RAM} e uma \ac{GPU} Tesla T4. Ademais, o processo de ajuste fino foi monitorado em tempo real utilizando o \textit{TensorBoard}, com relatório gerado durante as épocas de treinamento.

\begin{table}[ht!]
\centering
\caption{Detalhes de Treinamento dos Modelos no \textit{DeepfakeBenchmark}}
\scriptsize 
\begin{tabular}{|c|c|c|c|c|}
\hline
\rowcolor{lightgray}
\textbf{Modelo}          & \textbf{\textit{Backbone}} & \textbf{Classes} & \textbf{Conjunto} & \textbf{Aumentação?} \\ \hline
\textbf{Xception}        & Xception          & 2   & DeepSpeak         & Sim                  \\ \hline
\textbf{EfficientNet-B4} & EfficientNet-B4   & 2   & DeepSpeak         & Sim                  \\ \hline
\textbf{Meso4Inception}  & Meso4Inception    & 2   & DeepSpeak         & Não                  \\ \hline
\textbf{SPSL}            & Xception          & 2   & DeepSpeak         & Sim                  \\ \hline
\textbf{UCF}             & Xception          & 2   & DeepSpeak         & Sim                  \\ \hline
\end{tabular}
\label{tab:comparison_model_DFB}
\end{table}

\section{Experimentos Finais}
Os experimentos finais foram conduzidos com o objetivo de avaliar o desempenho dos modelos treinados e ajustados. Para isso, foi realizado um processo de predição utilizando o conjunto de dados reservado para testes, com foco em duas principais vertentes: a avaliação do desempenho do \textit{\ac{GenConViT}} de forma isolada e a comparação com os demais modelos presentes no \textit{DeepfakeBenchmark}. Ademais, apesar da possibilidade de concatenação dos resultados das redes presentes no \textit{\ac{GenConViT}} para predição, as redes foram avaliadas separadamente.

Primeiramente, foi avaliado o desempenho do \textit{\ac{GenConViT}} tanto no conjunto de dados \textit{WildDeepfake} quanto no \textit{DeepSpeak}. Também, foi realizada uma comparação detalhada entre o desempenho do \textit{\ac{GenConViT}} e os demais modelos utilizados no \textit{DeepfakeBenchmark}, especificamente para o conjunto de testes do \textit{DeepSpeak}. Essa comparação teve como objetivo analisar a eficácia do \textit{\ac{GenConViT}} em relação a outras arquiteturas. As métricas utilizadas são as mesmas citadas na Seção \ref{sec:exp_preliminar}.
\chapter{Resultados}
\label{resultados}

Neste Capítulo são apresentados os resultados dos experimentos realizados com o modelo \textit{GenConViT} e modelos do \textit{DeepfakeBenchmark}, destacando o desempenho na detecção de \ac{DF}. O objetivo é avaliar o desempenho do principal modelo de forma isolada e compará-lo com outras abordagens. Não obstante, os modelos apresentados passaram pelo processo de ajuste fino visando melhorar as métricas obtidas nos resultados finais. Para facilitar a compreensão dos resultados, é fundamental relembrar os conceitos da Seção \ref{metricas} acerca das principais métricas de avaliação utilizadas e sua importância no contexto de detecção de \ac{DF}. 

Dessa forma, a acurácia representa a proporção de previsões corretas em relação ao total de amostras analisadas, refletindo a capacidade do modelo de identificar corretamente vídeos reais e falsos. No entanto, essa métrica pode ser insuficiente quando há um desequilíbrio entre as classes (por exemplo, mais vídeos reais do que falsos). Todavia, os processos citados no Capítulo \ref{desenvolvimento} lidaram com essa problemática.

A precisão mede a proporção de vídeos classificados como falsos que são realmente \ac{DF}. Ou seja, a precisão avalia quantos dos vídeos rotulados como \ac{DF} são de fato falsos. Se a precisão for baixa, o modelo estará rotulando muitos vídeos reais como falsos, resultando em um número elevado de falsos positivos.

O \textit{recall} avalia a capacidade do modelo de detectar corretamente todos os \ac{DF} presentes no conjunto de dados. Essa métrica mede quantos dos vídeos falsos foram corretamente identificados pelo modelo. Um \textit{recall} baixo indica que o modelo está falhando em detectar diversos vídeos falsos, resultando em muitos falsos negativos, ou seja, vídeos falsos sendo classificados como verdadeiros.

O \textit{F1-score} é a média harmônica entre a precisão e o \textit{recall}. Essa métrica é particularmente útil quando há um desequilíbrio entre as classes, pois considera tanto a capacidade do modelo de identificar \ac{DF} quanto sua habilidade de evitar falsos positivos. Um \textit{F1-score} elevado indica que o modelo está equilibrando bem esses dois aspectos.

Por fim, a métrica \ac{AUC} - \ac{ROC} avalia a capacidade do modelo de distinguir entre vídeos reais e falsos ao variar os limiares de decisão. Uma \ac{AUC} próxima de 1 indica que o modelo possui excelente desempenho na separação entre amostras reais e falsas, enquanto uma \ac{AUC} de 0,5 reflete um desempenho aleatório.

\section{Resultados Iniciais}
Nesta Seção são apresentados os primeiros resultados obtidos durante os experimentos com o modelo \textit{GenConViT} em diferentes conjuntos de dados, especificamente nas bases \textit{DeepSpeak} e \textit{WildDeepfake}. Os resultados incluem uma análise detalhada do desempenho do modelo em termos de acurácia, precisão, \textit{recall}, \textit{F1-score} e \ac{AUC} - \ac{ROC}. São discutidos os métodos de falsificação da base \textit{DeepSpeak} que geraram mais falsos negativos e o impacto de variações de parâmetros, como a quantidade de quadros por vídeo utilizada durante a inferência. Também é avaliado como essas mudanças influenciam as métricas de desempenho do modelo.

Além do mais, são apresentados comparativos entre o \textit{GenConViT} e os modelos do \textit{DeepfakeBenchmark}, enfatizando quais abordagens foram mais eficientes no conjunto de dados \textit{DeepSpeak}. Esses resultados iniciais servem como base para ajustes futuros, buscando uma melhoria contínua na capacidade de detecção de \ac{DF}.

\subsection{Avaliação Inicial do \textit{GenConViT}}
\begin{figure}[ht!]
    \centering
    \includegraphics[width=0.7\textwidth]{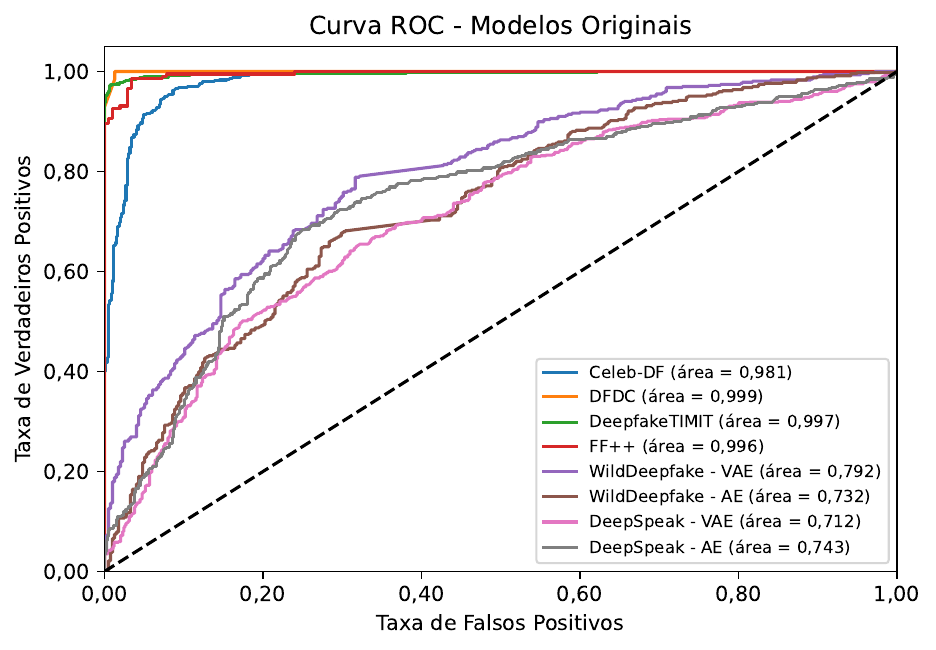}
    \caption{Desempenho do \textit{GenConViT} medido pela Curva ROC em múltiplas bases de dados, destacando a eficácia do modelo na distinção entre amostras reais e falsas.}
    \label{fig:genconvit_original_roc}
\end{figure}

No trabalho original do \textit{GenConViT}, \cite{wodajo2023deepfake} forneceu os resultados para as bases \textit{Celeb-DF (v2), DFDC, DeepfakeTIMIT} e \textit{FF++}, permitindo uma comparação direta entre os resultados obtidos no estado original do modelo e seu desempenho frente a novos conjuntos de dados. Pode-se observar que o modelo apresentou um desempenho elevado nas bases tradicionais com acurácias superiores a 90\%, conforme a Tabela \ref{tab:resultados_modelos_originais}. A \ac{AUC} também foi alta nesses conjuntos, variando de 0,981 a 0,999, o que indica uma excelente capacidade de separação entre amostras reais e falsas. O gráfico apresentado na Figura \ref{fig:genconvit_original_roc} ilustra as \ac{AUC} obtidas.

No entanto, ao testar o modelo em novos conjuntos de dados, houve uma queda no desempenho, especialmente quando se observa a diferença entre as Redes A -\textit{\ac{AE}} e B - \textit{\ac{VAE}}. Para a base \textit{WildDeepfake}, a Rede A obteve uma \ac{AUC} de 0,732, enquanto a Rede B alcançou \textbf{0,792}, com acurácias de 64,31\% e \textbf{69,64\%}, respectivamente. Isso indica que, apesar de ambos os métodos enfrentarem dificuldades em generalizar para falsificações mais complexas, a Rede B (\textit{VAE}) demonstrou um desempenho ligeiramente superior em termos de capacidade de separação e acurácia.

Para a base \textit{DeepSpeak}, o desempenho seguiu uma tendência similar. A Rede A obteve uma \ac{AUC} de \textbf{0,743} e uma acurácia de \textbf{68,48\%}, enquanto a Rede B apresentou uma \ac{AUC} de 0,712 e uma acurácia de 63,52\%. Embora a diferença de desempenho entre as redes não seja tão acentuada quanto no \textit{WildDeepfake}, a Rede A superou a Rede B nas detecções iniciais.

Esses resultados indicam que o \textit{GenConViT}, apesar de apresentar uma boa acurácia e \ac{AUC} em bases de dados estabelecidas, enfrenta desafios ao generalizar para novas bases de dados mais complexas e realistas. A diferença de desempenho entre a Rede A e a Rede B também sugere que a escolha da arquitetura tem um impacto importante na capacidade do modelo de lidar com diferentes tipos de falsificações, sendo que a Rede B foi mais eficaz no \textit{WildDeepfake}, enquanto a Rede A teve um melhor desempenho no \textit{DeepSpeak}. Para mais, a acurácia real permite notar que ambas as redes enfrentaram dificuldades em classificar amostrar reais como autênticas.

\begin{table}[ht!]
\centering
\caption{Resumo das métricas de avaliação do \textit{GenConViT} nos experimentos iniciais, comparando os resultados obtidos em diferentes conjuntos de dados.}
\label{tab:resultados_modelos_originais}
\resizebox{\textwidth}{!}{%
\begin{tabular}{|l|c|c|c|c|c|c|c|}
\hline
\rowcolor{lightgray}
\textbf{Base de Dados}            & \textbf{Acurácia (\%)} & \textbf{Acurácia Real (\%)} & \textbf{Acurácia Fake (\%)} & \textbf{ROC AUC} & \textbf{F1 Score} & \textbf{Precisão} & \textbf{Recall} \\ \hline
Celeb-DF                    & 90,95             & 83,05                  & 98,82                  & 0,981            & 0,916             & 0,854             & 0,988           \\ \hline
DFDC                        & 98,50             & 98,70                  & 98,45                  & 0,999            & 0,991             & 0,997             & 0,985           \\ \hline
DeepfakeTIMIT               & 97,02             & 96,35                  & 98,44                  & 0,997            & 0,955             & 0,928             & 0,984           \\ \hline
FF++                        & 97,05             & 95,59                  & 98,52                  & 0,996            & 0,971             & 0,957             & 0,985           \\ \hline
WildDeepfake - VAE        & \textbf{69,64}             & 54,94                  & 82,05                  & \textbf{0,792}            & 0,747             & 0,692             & 0,812           \\ \hline
WildDeepfake - AE         & 64,31             & 56,96                  & 70,51                  & 0,732            & 0,685             & 0,669             & 0,703           \\ \hline
DeepSpeak - VAE           & 63,52             & 88,32                  & 38,72                  & 0,712            & 0,504             & 0,755             & 0,378           \\ \hline
DeepSpeak - AE            & \textbf{68,48}             & 82,07                  & 54,89                  & \textbf{0,743}            & 0,627             & 0,749             & 0,539           \\ \hline
\end{tabular}%
 }
\end{table}

A Tabela \ref{tab:resultados_quadros} mostra o desempenho da Rede \textit{\ac{VAE}} no conjunto \textit{WildDeepfake} ao usar diferentes quantidades de quadros (10, 15 e 24). O melhor resultado foi obtido com 15 quadros, com uma acurácia de \textbf{69,64\%} e uma \ac{AUC} de \textbf{0,792}. Usar 10 ou 24 quadros resultou em um desempenho ligeiramente inferior, sugerindo que 15 quadros é o número aceitável para este conjunto de dados.

\begin{table}[ht!]
\centering
\caption{Comparação de desempenho com diferentes quantidades de quadros no \textit{WildDeepfake}.}
\label{tab:resultados_quadros}
\resizebox{\textwidth}{!}{%
\begin{tabular}{|l|c|c|c|c|c|c|c|c|l|}
\hline
\rowcolor{lightgray}
\textbf{Base de Dados} & \textbf{Rede} & \textbf{Quadros} & \textbf{Acurácia (\%)} & \textbf{Acurácia Real (\%)} & \textbf{Acurácia Fake (\%)} & \textbf{ROC AUC} & \textbf{F1 Score} & \textbf{Precisão} & \textbf{Recall} \\ \hline
WildDeepfake           & VAE           & 10f             & 68,60             & 52,91                  & 81,84                  & 0,785            & 0,742             & 0,682             & 0,814           \\ \hline
WildDeepfake           & VAE           & 15f             & \textbf{69,64}             & 54,94                  & 82,05                  & \textbf{0,792}            & 0,747             & 0,692             & 0,812           \\ \hline
WildDeepfake           & VAE           & 24f             & 68,60             & 54,18                  & 80,77                  & 0,786            & 0,743             & 0,687             & 0,808           \\ \hline
\end{tabular}%
}
\end{table}

\begin{figure}[ht!]
    \centering
    \includegraphics[width=0.8\textwidth]{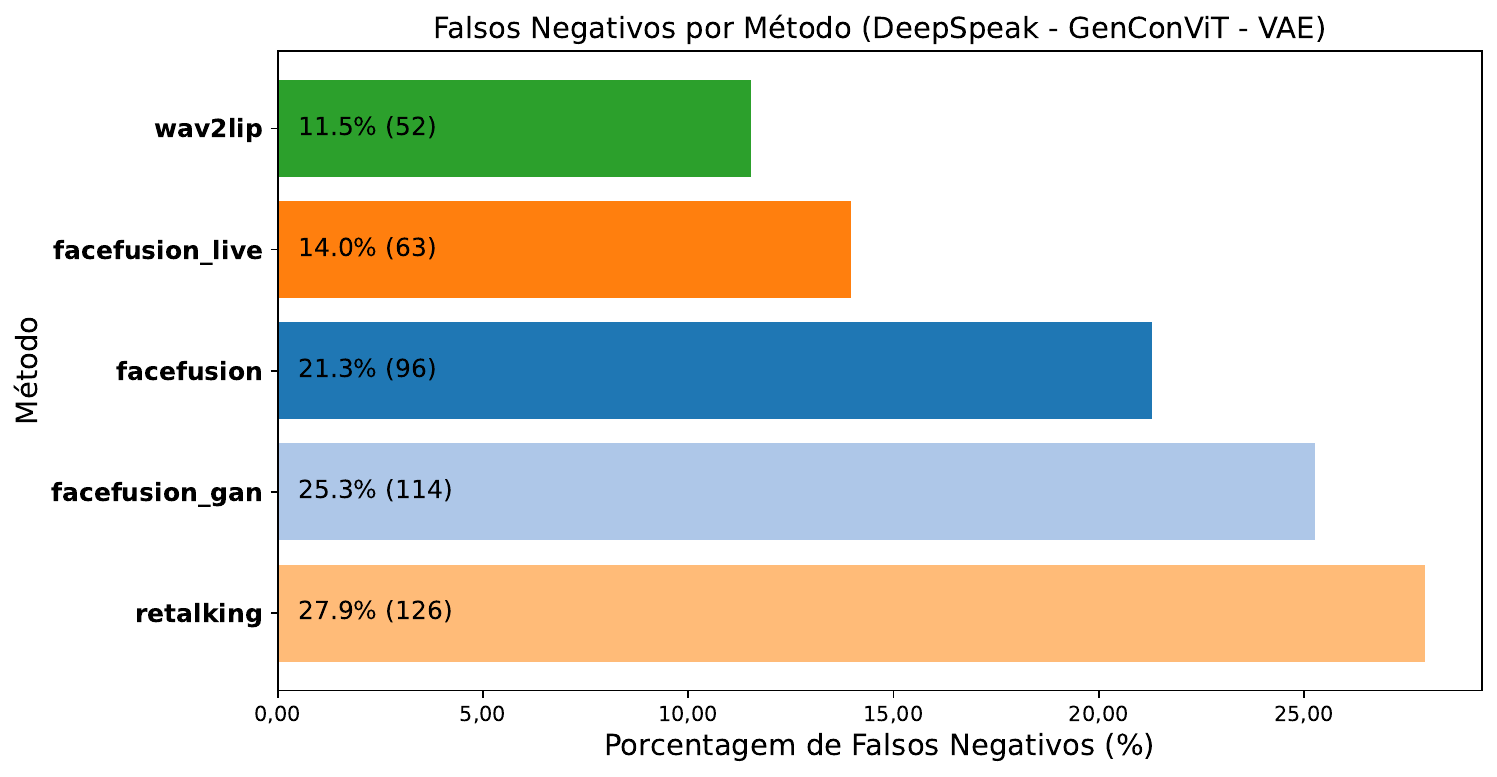}
    \caption{Distribuição de Falsos Negativos na Rede VAE para base \textit{DeepSpeak}.}
    \label{fig:distri_vae_deepspeak_false}
\end{figure}

\begin{figure}[ht!]
    \centering
    \includegraphics[width=0.8\textwidth]{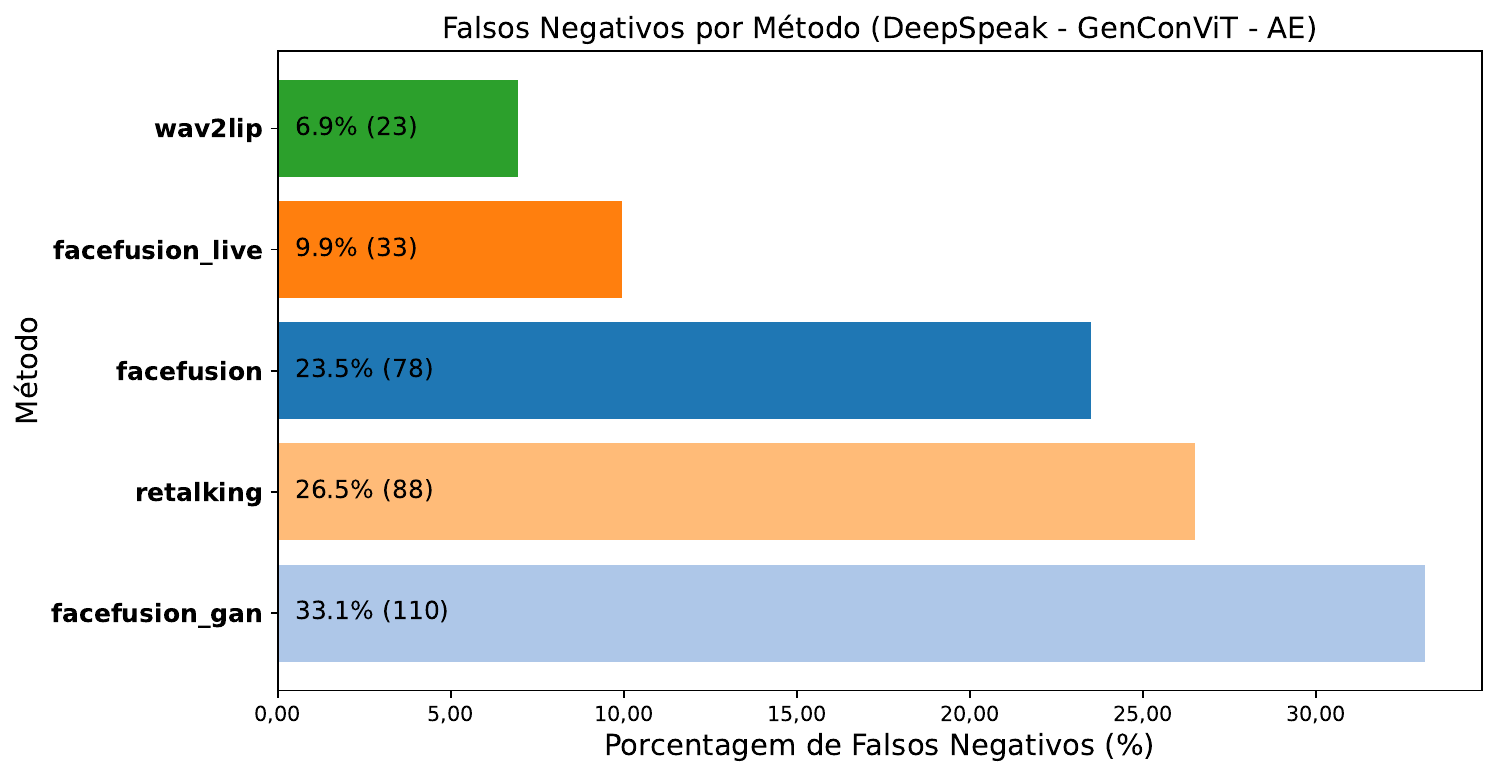}
    \caption{Distribuição de Falsos Negativos na Rede AE para base \textit{DeepSpeak}.}
    \label{fig:distri_ae_deepspeak_false}
\end{figure}

Um aspecto importante analisado foi a distribuição de falsos negativos, que no contexto de \ac{DF} representa o erro de classificar um vídeo falso como verdadeiro. Os resultados presentes nos gráficos das Figuras \ref{fig:distri_vae_deepspeak_false} e \ref{fig:distri_ae_deepspeak_false} mostraram que, na arquitetura \ac{VAE} do modelo \textit{GenConViT}, o método \textit{retalking} foi o maior responsável pelos falsos negativos, seguido por \textit{facefusion\_gan}, enquanto outros métodos, como \textit{facefusion} e \textit{wav2lip}, tiveram menor impacto. Na arquitetura \ac{AE}, a situação foi inversa, com \textit{facefusion\_gan} contribuindo mais para os falsos negativos, seguido por \textit{retalking}. 

\subsection{Comparação Inicial com Outros Modelos}
Após a avaliação inicial do modelo \textit{\ac{GenConViT}}, faz-se necessária a comparação com outros modelos de detecção de \ac{DF}, a fim de verificar se o desempenho observado está alinhado com o estado da arte ou se há oportunidades para melhorias. Para essa análise, foram selecionados os modelos \textit{Xception}, \textit{EfficientNetB4}, \textit{Meso4Inception}, \textit{SPSL} e \textit{UCF}, conhecidos pela alta performance na tarefa de detecção de falsificações.

\begin{figure}[ht!]
    \centering
    \includegraphics[width=0.8\textwidth]{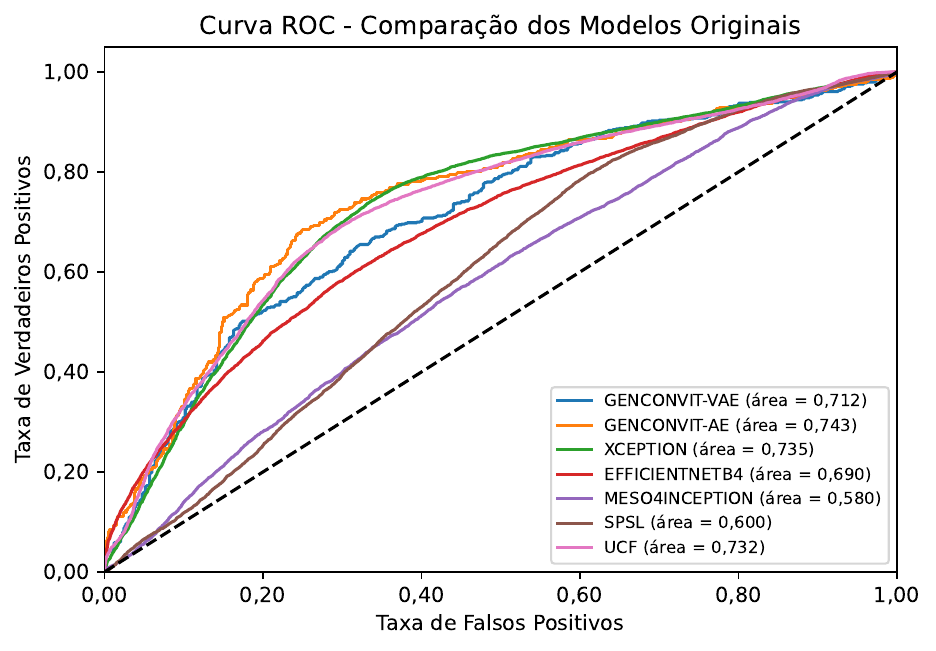}
    \caption{Desempenho do \textit{GenConViT} e modelos do \textit{Benchmark} Medido pela Curva ROC no \textit{DeepSpeak}.}
    \label{fig:curva_roc_benchmark}
\end{figure}

Como apresentado na Tabela \ref{tab:comparacao_modelos}, o modelo \textit{Xception} alcançou a melhor acurácia geral (\textbf{70,12\%}), com uma \ac{AUC} de \textbf{0,735}, conforme a Figura \ref{fig:curva_roc_benchmark} e boa proporção para acurácia \textit{real} e \textit{fake}. Esse resultado reflete uma considerável capacidade de separação entre amostras reais e falsas. Contudo, é importante notar que o \textit{\ac{GenConViT}} com arquitetura \ac{AE} também obteve resultados competitivos, apresentando uma \ac{AUC} superior de \textbf{0,743} e uma acurácia de 68,48\%. A versão da Rede \ac{VAE}, por outro lado, em termos de acurácia, alcançou 63,52\% com um maior desbalanço entre a acurácia para as classes, mas obteve uma \ac{AUC} de 0,712.

\begin{table}[ht!]
\centering
\caption{Métricas obtidas pelo \textit{GenConViT} e modelos do \textit{Benchmark} para o conjunto \textit{DeepSpeak}.}
\label{tab:comparacao_modelos}
\resizebox{\textwidth}{!}{%
\begin{tabular}{|l|c|c|c|c|c|c|c|}
\hline
\rowcolor{lightgray}
\textbf{Arquitetura} & \textbf{Acurácia (\%)} & \textbf{Acurácia Real (\%)} & \textbf{Acurácia Fake (\%)} & \textbf{ROC AUC} & \textbf{F1 Score} & \textbf{Precisão} & \textbf{Recall} \\ \hline
GevConViT-VAE        & 63,52             & 88,32                  & 38,72                  & 0,712            & 0,504             & 0,755             & 0,378           \\ \hline
GevConViT-AE         & 68,48             & 82,07                  & 54,89                  & \textbf{0,743}            & 0,627             & 0,749             & 0,539           \\ \hline
Xception             & \textbf{70,12}             & 67,74                  & 72,54                  & 0,735            & 0,707             & 0,690             & 0,725           \\ \hline
EfficientNetB4       & 63,81             & 60,96                  & 66,70                  & 0,690            & 0,647             & 0,628             & 0,667           \\ \hline
Meso4Inception       & 52,89             & 10,80                  & \textbf{95,48}                  & 0,580            & 0,668             & 0,514             & 0,955           \\ \hline
SPSL                 & 58,53             & 32,73                  & 84,62                  & 0,600            & 0,670             & 0,554             & \textbf{0,846}           \\ \hline
UCF                  & 69,45             & 72,35                  & 66,52                  & 0,732            & 0,684             & 0,704             & 0,665           \\ \hline
\end{tabular}%
}
\end{table}

Outro ponto de destaque é o comportamento do modelo \textit{Meso4Inception}, que apresentou a menor acurácia geral (52,89\%), mas uma alta precisão (\textbf{95,48\%}) para classificar amostras falsas. Esse contraste sugere que o modelo tem maior facilidade de detectar falsificações, resultando em poucos falsos positivos, mas falhando em detectar grande parte dos vídeos reais, como evidenciado por sua baixa acurácia para vídeos reais (10,80\%).

O modelo \textit{EfficientNetB4}, com uma acurácia de 63,81\% e \ac{AUC} de 0,690, teve um desempenho semelhante ao do \textit{GenConViT-VAE}, embora tenha se mostrado menos eficiente na classificação de vídeos reais. A acurácia real para o \textit{EfficientNetB4} foi de apenas 60,96\%, enquanto o \textit{GenConViT-VAE} alcançou 88,32\%.

Já o modelo \textit{SPSL} apresentou um desempenho moderado em quase todas as métricas, com destaque para seu \textit{recall} de \textbf{0,846}, o que indica que ele conseguiu detectar a maior parte dos vídeos falsos, embora com um custo em termos de precisão (0,554). Por outro lado, o modelo \textit{UCF} obteve um desempenho mais equilibrado, com uma acurácia de 69,45\%, e \ac{AUC} de 0,732, semelhante ao desempenho do \textit{Xception}, sugerindo que ele pode ser uma opção robusta para tarefas de detecção de \ac{DF}.

Ao comparar esses modelos com o \textit{GenConViT}, observa-se que, embora o \textit{GenConViT-AE} não tenha alcançado a melhor acurácia geral, ele teve um desempenho competitivo, superando a maioria dos modelos testados. Esse resultado reforça que o \textit{GenConViT}, especialmente em sua arquitetura \ac{AE}, tem um forte potencial para melhorar seu desempenho com ajustes mais refinados.

\section{Resultados Finais}
Nesta Seção, são apresentados os resultados obtidos após o ajuste fino do modelo \textit{GenConViT} para os conjuntos de dados \textit{WildDeepfake} e \textit{DeepSpeak}, incluindo uma análise detalhada do desempenho das redes \ac{AE} e \ac{VAE} e tempo de execução. Serão discutidas as diferenças de resultados para diferentes épocas de treinamento. Além disso, os resultados do \textit{GenConViT} ajustado serão confrontados com modelos do \textit{DeepfakeBenchmark} para verificar sua competitividade. Também, serão comparados os tempos de predição utilizando \ac{CPU} e \ac{GPU}, evidenciando as vantagens e limitações de cada abordagem. Por fim, a importância da aumentação de dados será abordada, destacando seu impacto na melhoria das métricas.

\subsection{Avaliação Final do \textit{GenConViT}}
\label{avaliacao_final}
\begin{figure}[ht!]
    \centering
    \includegraphics[width=0.7\textwidth]{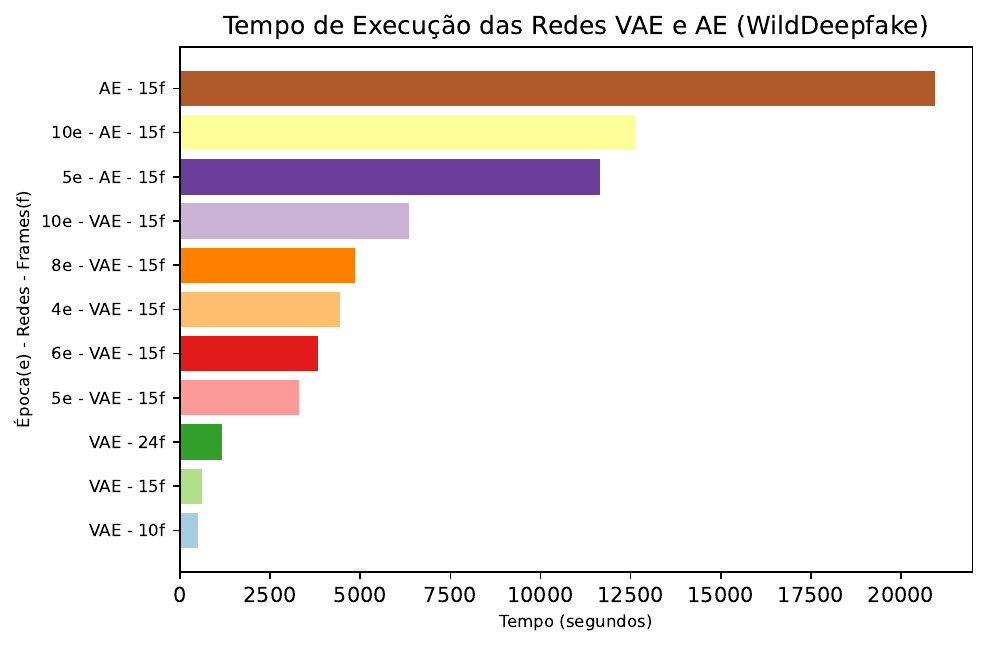}
    \caption{Tempo gasto para predição com modelo ajustado para diferentes épocas no \textit{WildDeepfake}.}
    \label{fig:exec_time_wild}
\end{figure}

\begin{figure}[ht!]
    \centering
    \includegraphics[width=0.7\textwidth]{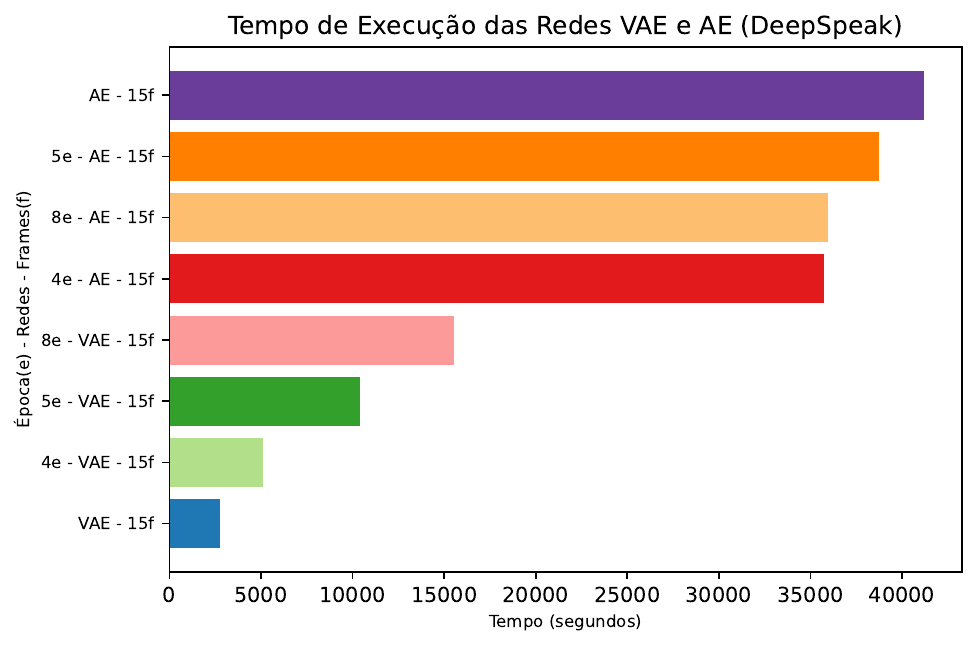}
    \caption{Tempo gasto para predição com modelo ajustado para diferentes épocas no \textit{DeepSpeak}.}
    \label{fig:exec_time_speak}
\end{figure}

A princípio, as predições foram realizadas utilizando uma \ac{CPU}, e o tempo total gasto para predição de cada estado do modelo (definido pela quantidade de épocas) foi registrado. De forma geral, a Rede A - \ac{AE} apresentou um tempo de execução significativamente maior em comparação à Rede B - \ac{VAE} para ambos os conjuntos de dados, conforme demonstrado nos gráficos das Figuras \ref{fig:exec_time_wild} e \ref{fig:exec_time_speak}.

Outro aspecto observado foi o tamanho dos arquivos de pesos (\textit{checkpoint}) gerados após o treinamento do \textit{GenConViT}. Os arquivos salvos para a Rede \ac{AE} possuem aproximadamente 450MB, enquanto os da Rede \ac{VAE} têm 5,2GB. Ademais, a Rede \ac{VAE} é treinada de forma mais rápida para uma mesma quantidade de épocas e, concomitantemente, maior foi a carga de testes com esta. Vale salientar que as discrepâncias de tempo entre as redes já eram observadas antes do ajuste fino, e continuaram presentes mesmo após as otimizações realizadas. Isso sugere que a Rede \ac{VAE} possui uma arquitetura mais eficiente para operações rápidas demandando maior quantidade de \ac{RAM} para carregar os pesos, enquanto a Rede \ac{AE} pode demandar recursos computacionais e maior tempo de processamento.

\begin{figure}[ht!]
    \centering
    \includegraphics[width=0.7\textwidth]{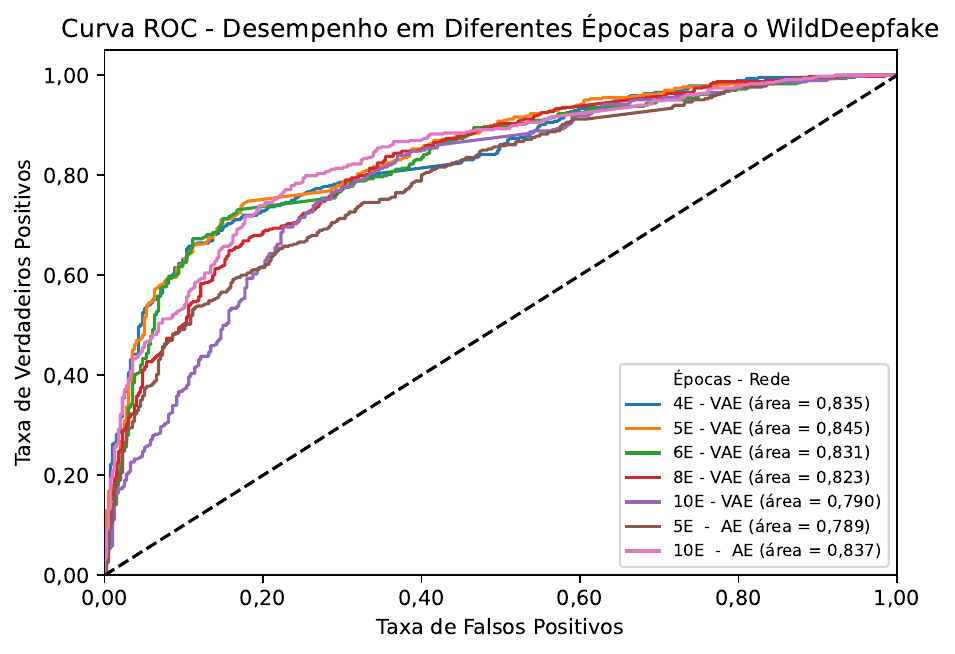}
    \caption{Desempenho do \textit{GenConViT} medido pelo Curva ROC no \textit{WildDeepfake} variando a quantidade de épocas.}
    \label{fig:roc_retrain_wild}
\end{figure}

Após o ajuste fino no conjunto \textit{WildDeepfake}, as redes A - \ac{AE} e B - \ac{VAE} foram testadas variando a quantidade de épocas de treinamento. Os resultados são apresentados na Tabela \ref{tab:genconvit_wilddeepfake}, destacando que a rede \ac{VAE} obteve uma acurácia máxima de \textbf{74,62\%} para 5 épocas, enquanto a \ac{ROC} - \ac{AUC} mais alta foi \textbf{0,845}, conforme o gráfico da Figura \ref{fig:roc_retrain_wild}. A rede \ac{AE} apresentou seu melhor desempenho com 10 épocas, atingindo uma \ac{ROC} - \ac{AUC} de 0,837 e uma acurácia de 65,47\%. Observa-se que a Rede \ac{VAE} demonstrou uma maior estabilidade e precisão em relação à Rede \ac{AE}. Contudo, houve uma relativa dificuldade do modelo para se adaptar ao \textit{WildDeepfake} em ambas as redes.

\begin{table}[ht!]
\centering
\caption{\textit{GenConViT} Reajustado no \textit{WildDeepfake}}
\resizebox{\textwidth}{!}{%
\begin{tabular}{|c|c|c|c|c|c|c|c|c|}
\hline
\rowcolor{lightgray}
\textbf{Rede} & \textbf{Época} & \textbf{Acurácia (\%)} & \textbf{Acurácia Real (\%)} & \textbf{Acurácia Fake (\%)} & \textbf{ROC AUC} & \textbf{F1 Score} & \textbf{Precisão} & \textbf{Recall} \\ \hline
VAE           & 4              & 72,42                   & 55,95                        & 86,32                        & 0,835           & 0,748             & 0,684             & 0,825           \\ \hline
VAE           & 5              & \textbf{74,62}                   & 70,13                        & 78,42                        & \textbf{0,845}           & 0,765             & 0,761             & 0,769           \\ \hline
VAE           & 6              & 74,39                   & 73,16                        & 75,43                        & 0,831           & 0,759             & 0,765             & 0,752           \\ \hline
VAE           & 8              & 65,24                   & 28,10                        & 96,58                        & 0,823           & 0,753             & 0,620             & 0,957           \\ \hline
VAE           & 10             & 69,41                   & 45,82                        & 89,32                        & 0,790           & 0,758             & 0,664             & 0,882           \\ \hline
AE            & 5              & 63,50                   & 27,59                        & 93,80                        & 0,789           & 0,736             & 0,608             & 0,934           \\ \hline
AE            & 10             & 65,47                   & 30,89                        & 94,66                        & 0,837           & 0,749             & 0,624             & 0,938           \\ \hline
\end{tabular}%
}
\label{tab:genconvit_wilddeepfake}
\end{table}

\begin{figure}[ht!]
    \centering
    \includegraphics[width=0.7\textwidth]{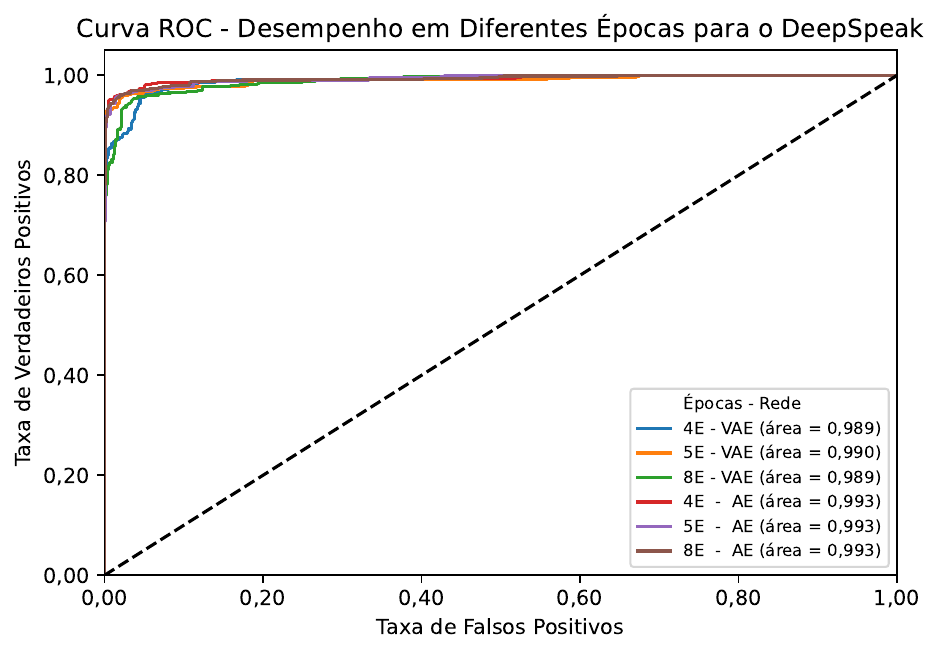}
    \caption{Desempenho do \textit{GenConViT} medido pelo Curva ROC no \textit{DeepSpeak} variando a quantidade de épocas.}
    \label{fig:roc_retrain_speak}
\end{figure}

Para o conjunto \textit{DeepSpeak}, os resultados pós-ajuste fino mostraram um desempenho superior em comparação ao \textit{WildDeepfake}, especialmente para a Rede \ac{AE}. Conforme apresentado na Tabela \ref{tab:genconvit_deepspeak}, a rede AE alcançou uma acurácia de \textbf{96,67\%} e \ac{ROC} - \ac{AUC} de \textbf{0,993} para 4 épocas, representando o melhor resultado entre todas as configurações testadas, conforme o gráfico da Figura \ref{fig:roc_retrain_speak}. A rede \ac{VAE}, embora tenha obtido resultados muito bons, apresentou uma acurácia máxima de 95,45\% para 4 épocas. Essa performance sugere que a arquitetura AE pode ser mais eficaz para o tipo de manipulação presente no \textit{DeepSpeak}, enquanto a \ac{VAE} mostrou uma leve vantagem na estabilidade de métricas como precisão e \textit{recall}.

\begin{table}[ht!]
\centering
\caption{\textit{GenConViT} Reajustado no \textit{DeepSpeak}}
\resizebox{\textwidth}{!}{%
\begin{tabular}{|c|c|c|c|c|c|c|c|c|}
\hline
\rowcolor{lightgray}
\textbf{Rede} & \textbf{Época} & \textbf{Acurácia (\%)} & \textbf{Acurácia Real (\%)} & \textbf{Acurácia Fake (\%)} & \textbf{ROC AUC} & \textbf{F1 Score} & \textbf{Precisão} & \textbf{Recall} \\ \hline
VAE           & 4              & 95,45                   & 93,21                        & 97,69                        & 0,989           & 0,950             & 0,931             & 0,970           \\ \hline
VAE           & 5              & 90,42                   & 82,07                        & 98,78                        & 0,990           & 0,909             & 0,844             & 0,985           \\ \hline
VAE           & 8              & 93,27                   & 88,72                        & 97,83                        & 0,989           & 0,929             & 0,887             & 0,974           \\ \hline
AE            & 4              & \textbf{96,67}                   & 95,11                        & 98,23                        & \textbf{0,993}           & 0,965             & 0,951             & 0,980           \\ \hline
AE            & 5              & 93,82                   & 88,86                        & 98,78                        & \textbf{0,993}           & 0,938             & 0,897             & 0,984           \\ \hline
AE            & 8              & 94,02                   & 89,27                        & 98,78                        & \textbf{0,993}           & 0,942             & 0,902             & 0,985           \\ \hline
\end{tabular}%
}
\label{tab:genconvit_deepspeak}
\end{table}

Os tempos de execução médios das redes VAE e ED para os conjuntos de dados após o ajuste fino são apresentados na Tabela \ref{tab:execution_time}. Para cada rede e configuração, o tempo total de execução e a quantidade de amostras foram utilizados para calcular o tempo médio por amostra, permitindo uma comparação clara entre as arquiteturas. Nesse cenário, a rede \ac{VAE} demonstrou maior eficiência, enquanto a rede \ac{AE} teve tempos mais elevados apesar de uma acurácia levemente superior. Vale lembrar que há a possibilidade de concatenar os resultados das redes conforme a arquitetura padrão do \textit{\ac{GenConViT}} e, concomitantemente, o tempo gasto com a predição será maior.

\begin{table}[ht!]
\centering
\caption{Tempo de execução médio e acurácia das redes \textit{GenConViT} para \textit{WildDeepfake} e \textit{DeepSpeak}}
\label{tab:execution_time}
\resizebox{\textwidth}{!}{%
\begin{tabular}{|c|c|c|c|c|c|c|}
\hline
\rowcolor{lightgray}
\textbf{Base de Dados}    & \textbf{Rede} & \textbf{Época (Ajuste)} & \textbf{Acurácia (\%)} & \textbf{Tempo Total (s)} & \textbf{Total de Amostras} & \textbf{Tempo Médio (s/amostra)} \\ \hline
WildDeepfake        & VAE           & 5              & 74,62                  & 3292                     & 863                       & 3,81                             \\ \hline
WildDeepfake        & AE            & 10             & 65,47                  & 11644                    & 863                       & 13,5                             \\ \hline
DeepSpeak           & VAE           & 4              & 95,45                  & 5097                     & 1472                       & 3,46                             \\ \hline
DeepSpeak           & AE            & 4              & 96,67                  & 35753                    & 1472                       & 24,29                            \\ \hline
\end{tabular}%
}
\end{table}

\subsection{Comparação Final do \textit{GenConViT} com Outros Modelos}
Para garantir uma comparação justa entre os modelos ajustados, foi utilizado o \textit{GenConViT} com 5 épocas, tanto para a arquitetura \ac{VAE} quanto para a \ac{AE}, apesar de essas configurações não terem sido as que apresentaram o melhor desempenho absoluto nos resultados da Subseção \ref{avaliacao_final}. Vale lembrar que o processo de treinamento dos modelos do \textit{Benchmark} foram monitorados com \textit{TensorBoard} e os relatórios gerados serão disponibilizados no repositório citado no Capítulo \ref{desenvolvimento}. As métricas detalhadas, conforme apresentadas na Tabela \ref{tab:comparison_metrics_benchmark_retrain}, mostram que o \textit{SPSL} obteve a melhor acurácia entre todos os modelos, com \textbf{93,96\%}, e uma ROC AUC de 0,983. Em contraste, o \textit{GenConViT - 5E - \ac{AE}} também apresentou um excelente desempenho, com uma acurácia de 93,82\% e a maior \ac{ROC} - \ac{AUC} da comparação, \textbf{0,993}. Já o \textit{GenConViT - 5E - \ac{VAE}} obteve uma boa acurácia de 90,42\% e uma \ac{ROC} - \ac{AUC} de 0,990.

\begin{table}[ht!]
\centering
\caption{Comparação de Métricas - Modelos Ajustados}
\label{tab:comparison_metrics_benchmark_retrain}
\resizebox{\textwidth}{!}{%
\begin{tabular}{|c|c|c|c|c|c|c|c|}
\hline
\rowcolor{lightgray}
\textbf{Arquitetura} & \textbf{Acurácia (\%)} & \textbf{Acurácia Real (\%)} & \textbf{Acurácia Fake (\%)} & \textbf{ROC AUC} & \textbf{F1 Score} & \textbf{Precisão} & \textbf{Recall} \\ \hline
GenConViT-5E-VAE     & 90,42                  & 82,07                        & 98,78                        & 0,990           & 0,909             & 0,844             & 0,985           \\ \hline
GenConViT-5E-AE      & 93,82                  & 88,86                        & 98,78                        & \textbf{0,993}           & 0,938             & 0,897             & 0,984           \\ \hline
Xception             & 87,90                  & 82,59                        & 93,28                        & 0,964           & 0,885             & 0,841             & 0,933           \\ \hline
EfficientNetB4       & 91,60                  & 92,98                        & 90,21                        & 0,975           & 0,914             & 0,927             & 0,902           \\ \hline
Meso4Inception       & 69,29                  & 92,64                        & 45,66                        & 0,812           & 0,596             & 0,860             & 0,457           \\ \hline
SPSL                 & \textbf{93,96}                  & 94,40                        & 93,51                        & 0,983           & 0,939             & 0,943             & 0,935           \\ \hline
UCF                  & 86,41                  & 80,04                        & 92,85                        & 0,950           & 0,872             & 0,821             & 0,929           \\ \hline
\end{tabular}%
}
\end{table}

Ao comparar esses resultados com outros modelos, observa-se que o \textit{Xception} e o \textit{EfficientNetB4} também apresentaram boas performances, com acurácias de 87,90\% e 91,60\%, respectivamente, e áreas sob a curva \ac{ROC} de 0,964 e 0,975. No entanto, o modelo \textit{Meso4Inception} teve um desempenho substancialmente inferior, com uma acurácia de apenas 69,29\% e uma \ac{ROC} - \ac{AUC} de 0,812. Esse resultado pode estar relacionado à falta de estratégias de aumentação de dados durante o treinamento do \textit{Meso4Inception}, o que limita sua capacidade de generalização e impacta negativamente a detecção de falsos positivos e falsos negativos, como evidenciado por sua baixa acurácia para amostras falsas (45,66\%).

\begin{figure}[ht!]
    \centering
    \includegraphics[width=0.7\textwidth]{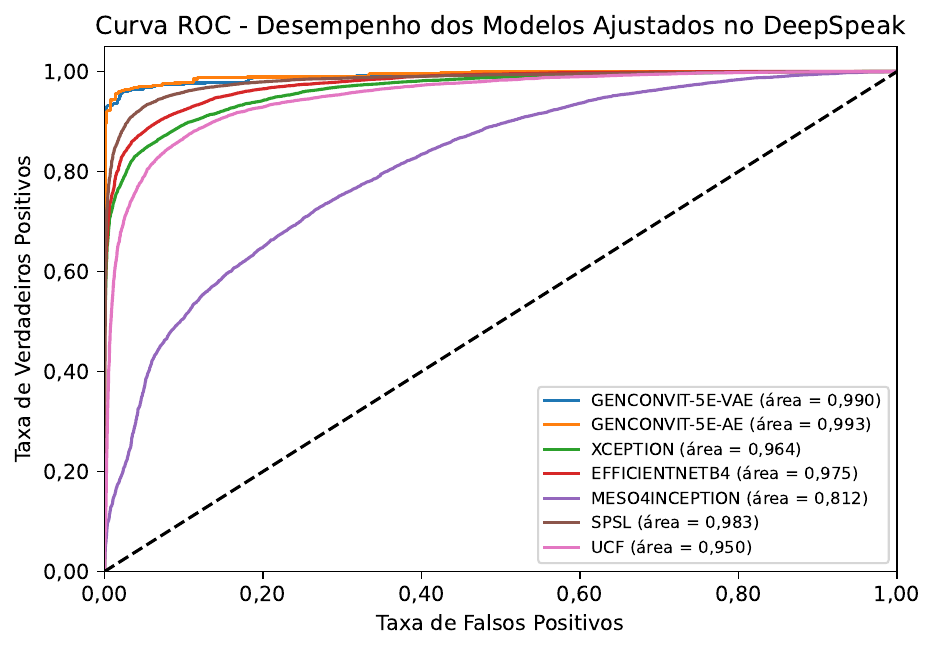}
    \caption{Desempenho dos modelos medido pelo curva ROC no \textit{DeepSpeak}.}
    \label{fig:roc_retrain_benchmark}
\end{figure}

Bem como, o gráfico da Curva ROC apresentado na Figura \ref{fig:roc_retrain_benchmark} reflete visualmente as diferenças entre os modelos. Ambas as redes do \textit{GenConViT} destacam-se com áreas sob a curva muito próximas de 1, evidenciando sua alta capacidade de separação entre amostras reais e falsas. Por outro lado, o \textit{Meso4Inception} apresentou uma curva muito mais próxima da linha de aleatoriedade, confirmando sua baixa eficiência.

\begin{table}[ht!]
\centering
\caption{Comparação de acurácia e AUC dos modelos antes e após o ajuste fino no \textit{DeepSpeak}}
\label{tab:comparacao_acuracia_auc}
\begin{tabular}{|l|c|c|c|c|}
\hline
\rowcolor{lightgray}
\textbf{Modelo} & \textbf{Acurácia (Pré)} & \textbf{Acurácia (Pós)} & \textbf{AUC (Pré)} & \textbf{AUC (Pós)} \\ \hline
GenConViT-VAE      & 63,52 & 90,42 & 0,712 & 0,990 \\ \hline
GenConViT-AE       & 68,48 & 93,82 & 0,743 & 0,993 \\ \hline
Xception           & 70,12 & 87,90 & 0,735 & 0,964 \\ \hline
EfficientNetB4     & 63,81 & 91,60 & 0,690 & 0,975 \\ \hline
Meso4Inception     & 52,89 & 69,29 & 0,580 & 0,812 \\ \hline
SPSL               & 58,53 & 93,96 & 0,600 & 0,983 \\ \hline
UCF                & 69,45 & 86,41 & 0,732 & 0,950 \\ \hline
\end{tabular}
\end{table}

Ademais, os experimentos também evidenciaram uma abrupta mudança na velocidade de predição com o uso de \ac{GPU} em vez de \ac{CPU}. Um exemplo notável foi o modelo \textit{UCF}, cujo tempo de predição das 1.472 amostras de teste caiu drasticamente de 8 horas e 31 minutos para apenas 3 minutos e 31 segundos. De forma similar, o \textit{GenConViT} também apresentou uma melhora significativa na velocidade de predição com a utilização de \ac{GPU}, destacando a importância de recursos de hardware robustos para o processamento eficiente e em larga escala na detecção de \ac{DF}.

Para finalizar, a Tabela \ref{tab:comparacao_acuracia_auc} apresenta uma síntese comparativa dos resultados de acurácia e AUC dos modelos avaliados antes e após o ajuste fino. É possível observar melhorias significativas nas métricas para todos os modelos após o refinamento, especialmente para o \textit{GenConViT}, que alcançou desempenho competitivo com o estado da arte na detecção de \ac{DF}.

\chapter{Conclusões e Trabalhos Futuros}
\label{conclusoes}

Este trabalho teve como objetivo principal comparar o estado da arte dos modelos de detecção de \ac{DF}, tanto sem quanto com ajuste fino, utilizando conjuntos de dados modernos e desafiadores como \textit{WildDeepfake} e \textit{DeepSpeak}. Através de uma metodologia estruturada, que envolveu a seleção de bases de dados, adaptação de arquiteturas, pré-processamento, realização de experimentos iniciais e finais, foi possível avaliar a eficácia do modelo \textit{\ac{GenConViT}} em comparação com outras arquiteturas presentes no \textit{DeepfakeBenchmark}.

Os resultados obtidos demonstraram que o \textit{\ac{GenConViT}} é altamente competitivo em relação às demais arquiteturas avaliadas, especialmente após a realização do ajuste fino. A Rede A (\ac{AE}) do \textit{\ac{GenConViT}} alcançou uma acurácia de 93,82\% e uma \ac{ROC} - \ac{AUC} de 0,993 no conjunto \textit{DeepSpeak}, superando modelos tradicionais como \textit{Xception} e \textit{EfficientNetB4} em métricas chave. Isso evidencia a eficácia do \textit{\ac{GenConViT}} na detecção de \ac{DF}, particularmente quando adaptado para conjuntos de dados mais complexos e realistas. Também, a personalização do modelo para cenários específicos aumentou significativamente sua capacidade de generalização, permitindo uma detecção mais precisa de \ac{DF} em diferentes contextos. No entanto, apesar do excelente desempenho em bases de dados tradicionais, o modelo enfrentou desafios ao generalizar para novos conjuntos como \textit{WildDeepfake}. Isso ressalta a importância de treinar modelos com dados diversificados para melhorar sua capacidade de detecção em cenários do mundo real, uma vez que os modelos presentes no \textit{benchmark} também foram positivamente impactados pelo ajuste fino.

Outro aspecto relevante observado foi a eficiência computacional dos modelos. A utilização de \ac{GPU}s mostrou-se essencial para acelerar o processo de treinamento e predição, reduzindo significativamente os tempos de execução. Adicionalmente, a comparação dos resultados entre as Redes A (\ac{AE}) e B (\ac{VAE}) do \textit{\ac{GenConViT}} revelou que a Rede B (\ac{VAE}) apresentou um desempenho ligeiramente superior em termos de estabilidade e precisão no conjunto \textit{WildDeepfake}, enquanto a Rede A (\ac{AE}) teve um melhor desempenho no conjunto \textit{DeepSpeak}. Essa diferença sugere que a escolha da arquitetura impacta significativamente a capacidade do modelo de lidar com diferentes tipos de falsificações. Nesse contexto, a utilização de arquiteturas multimodais surge como uma abordagem promissora, permitindo combinar diferentes modalidades de dados, como áudio, vídeo e texto, para capturar informações complementares e melhorar as capacidades dos modelos de detecção frente à variabilidade presente nos dados de diferentes bases.

Vale destacar que, embora redes neurais profundas sejam amplamente utilizadas para gerar falsificações sofisticadas, como na criação de \textit{Deepfakes}, elas também têm se mostrado ferramentas úteis no combate a essas manipulações. Um exemplo notável é o uso do \ac{VAE}, empregado em ambas as tarefas. No modelo \textit{\ac{GenConViT}}, a Rede B (\ac{VAE}) desempenhou um papel crucial na identificação de inconsistências em vídeos manipulados, utilizando sua capacidade de modelar distribuições no espaço latente para capturar padrões que diferenciam mídias reais de falsificadas. Todavia, as redes \ac{VAE} são amplamente reconhecidas como uma das principais arquiteturas de \ac{IA} generativa. Essa dualidade reflete a capacidade dessas redes de realizar diversas tarefas, atendendo a cenários variados com eficiência e adaptabilidade.

Para trabalhos futuros, sugere-se adaptar o \textit{\ac{GenConViT}} para integração ao \textit{DeepfakeBenchmark}, incorporando novos conjuntos de dados. Essa adaptação permitirá uma avaliação mais ampla e padronizada de seu desempenho em comparação com outros modelos disponíveis na plataforma. Além disso, recomenda-se obter e treinar os modelos do \textit{benchmark} utilizando o conjunto de dados original do \textit{WildDeepfake}. Essa abordagem possibilitará um pré-processamento padronizado, uma análise detalhada das características e técnicas de \ac{DF} presentes nesse conjunto, e a aplicação de diferentes arquiteturas para detecção. Por fim, sugere-se a exploração dos demais modelos disponíveis no \textit{benchmark}, ampliando o escopo de redes neurais analisadas e investigando novas abordagens que possam oferecer melhor desempenho ou eficiência na detecção de \ac{DF}.

Em considerações finais, a detecção de \ac{DF} é um campo em constante evolução, impulsionado pelos avanços das técnicas de geração de vídeos falsificados. Este trabalho contribuiu para o entendimento e aprimoramento das capacidades dos modelos de aprendizado profundo na identificação dessas manipulações, demonstrando que arquiteturas híbridas como o \textit{\ac{GenConViT}} podem oferecer desempenho competitivo e robusto. Todavia, a pesquisa também revelou a importância de continuamente atualizar e treinar modelos com dados diversificados e realistas para acompanhar o ritmo acelerado das inovações em \ac{DF}. Através das conclusões e dos trabalhos futuros propostos, espera-se que futuras pesquisas possam expandir ainda mais as fronteiras da detecção de \ac{DF}, desenvolvendo soluções mais eficazes e adaptáveis para proteger a autenticidade da informação em um mundo digital cada vez mais complexo.

\bibliographystyle{apalike2_ptBR} 

\bibliography{referencias}


\apendices 

\anexos 

\end{document}